\documentclass[11pt]{article}

\usepackage[preprint]{acl}
\usepackage[dvipsnames, table]{xcolor}
\usepackage{makecell}
\usepackage{array}
\definecolor{MyGreen}{HTML}{00B050}
\definecolor{srcColor}{HTML}{B03A2E} 
\definecolor{tgtColor}{HTML}{2874A6}
\usepackage{colortbl}

\usepackage{times}
\usepackage{latexsym}
\usepackage{CJKutf8}
\usepackage{amsfonts}
\usepackage{amssymb}
\usepackage{geometry}
\usepackage{hyperref}
\usepackage[T1]{fontenc}

\usepackage[utf8]{inputenc}
\usepackage{amsmath}
\usepackage{microtype}
\usepackage{booktabs}
\usepackage{mathtools}
\usepackage{multirow}
\usepackage{multicol}
\usepackage{nicematrix}

\usepackage{inconsolata}

\usepackage{graphicx}
\usepackage{enumitem}
\usepackage{threeparttable}
\newcommand{\ignore}[1]{}

\title{STAR : Sentence Translation Alignment Rate for Document-to-Document Machine Translation}

\author{
\textbf{Yichen Dong}$^{1}$\thanks{Equal contribution.} \quad
\textbf{Hao Wang}$^{2}$\footnotemark[1] \quad
\textbf{Junhui Li}$^{1}$\thanks{Corresponding author.} \\
\textbf{Linlong Xu}$^{2}$ \quad
\textbf{Longyue Wang}$^{2}$ \quad
\textbf{Weihua Luo}$^{2}$ \\
$^{1}$School of Computer Science and Technology, Soochow University, Suzhou, China \\
$^{2}$Alibaba Group, Hangzhou, China \\
\texttt{dong\_yichen@foxmail.com, huaiyu.wh@alibaba-inc.com, jhli@suda.edu.cn}
}

\begin{document}
\maketitle


\begin{abstract}
Large Language Models (LLMs) have enabled a shift from sentence-level to document-to-document (Doc2Doc) machine translation, promising improved global coherence. However, document-to-document generation in a single pass frequently suffers from structural misalignment, manifesting as sentence omissions or hallucinations that violate the core requirement of source-target correspondence. To address this, we introduce \textbf{S}entence \textbf{T}ranslation \textbf{A}lignment \textbf{R}ate (\textbf{STAR}), an auxiliary metric that explicitly quantifies sentence-level structural fidelity. Building on this, we propose \textbf{STAR}-masked \textbf{P}reference \textbf{O}ptimization (\textbf{StarPO}), a framework that ranks document-level hypotheses by structural quality and utilizes a dynamic alignment mask to focus optimization on misaligned segments. Experimental results across news and literary domains demonstrate that StarPO significantly enhances translation quality and structural integrity. Notably, StarPO allows compact models to surpass the performance of massive proprietary systems like GPT-4o while maintaining superior token efficiency.\footnote{Our code is available at \href{https://github.com/1078966865/STAR}{\nolinkurl{github.com/1078966865/STAR}}.}
\end{abstract}

\section{Introduction}

Recent advances in large language models (LLMs) has significantly advanced document-level machine translation (DocMT)~\cite{wang-etal-2023-document-level,wang-etal-2025-delta,karpinska-iyyer-2023-large}. With long-context modeling and strong generative capabilities, LLMs make it increasingly feasible to move beyond sentence-to-sentence (Sent2Sent) translation to chunk-to-chunk (Chunk2Chunk) and document-to-document (Doc2Doc) translation, where an entire source document is translated in a single pass. Despite this promise, Doc2Doc translation often exhibits sentence-level structural failures, such as sentence omissions or hallucinations, that violate the core requirement of sentence-level correspondence between source and target documents~\cite{dong-etal-2025-two}. In this work, we aim to improve Doc2Doc translation by addressing sentence-level structural misalignments, so that better alignment directly contributes to higher translation quality.

Most existing DocMT approaches implicitly avoid structural misalignment by adopting Sent2Sent or Chunk2Chunk paradigms, where translation is still performed at the individual sentence level given surrounding context~\cite{wu-etal-2024-adapting,hu-etal-2025-source,li-etal-2026-cross} or sliding chunks~\cite{liu-etal-2025-improving,li-etal-2026-enhancing}. While these localized paradigms successfully enforce sentence or block correspondence, they inherently limit global text planning and incur substantial computational costs, as source segments are repeatedly encoded across overlapping contexts.\footnote{See Appendix~\ref{apdx:efficiency} for additional discussions.}

\begin{table}[t]
\centering
\small
\setlength{\tabcolsep}{4.5pt}
\begin{NiceTabular}{lcccc}
\toprule
\Block{2-1}{\textbf{System}} & \textbf{Ideal} & \multicolumn{3}{c}{\textbf{Structural Deviations}} \\
\cmidrule(lr){2-2} \cmidrule(lr){3-5}
 & \textbf{1-to-1} & \textbf{1-to-0} & \textbf{0-to-1} & \textbf{Other} \\
\midrule
\rowcolor{gray!15} 
\Block[l]{1-5}{\textit{Doc2Doc translation in a single pass}} \\
LLaMA-3.1-8B & 92.59 & 2.08 & 1.17 & 4.16 \\
Qwen-2.5-7B & 95.35 & 1.91 & 1.31 & 1.43 \\
Qwen-3-4B & 94.72 & 0.72 & 0.33 & 4.23 \\
Deepseek-R1 & 95.03 & 4.85 & 0.03 & 0.09 \\
GPT-4o & 92.91 & 2.25 & 2.89 & 1.95 \\
\midrule
\rowcolor{gray!15} 
\Block[l]{1-5}{\textit{Chunk2Chunk prompting} LLaMA-3.1 \textit{w. sentence  boundary}} \\
MixSFT~\cite{li-etal-2026-enhancing} & 96.78 & 0.57 & 2.08 & 0.57 \\
KFMT~\cite{liu-etal-2025-improving} & 95.48 & 1.89 & 2.34 & 0.29 \\
\midrule
Ours (on Qwen2.5-7B) & 98.43 & 0.68 & 0.00  & 0.89  \\
\bottomrule
\end{NiceTabular}
\caption{Alignment distribution ( in \% ) for Zh $\Rightarrow$ En on News-Commentary, computed by Gemini-2.5-Flash.}
\label{tab:alignment_stats}
\end{table}

Ideally, Doc2Doc translation preserves a clear correspondence between source and target sentences. In practice, however, model outputs often contain 1-to-0 alignments (omissions), 0-to-1 alignments (hallucinations) or even generation collapse (e.g., repetitive loops). Are non-1-to-1 alignments, where $N$ source sentences correspond to $M$ ($M \neq N$) target sentences, inherently unacceptable? Intuitively, cross-lingual syntax often justifies merging, splitting, or reordering for fluency, ensuring the output feels more native and adheres to the idiomatic conventions of the target language. Therefore, robust Doc2Doc systems must therefore distinguish such legitimate restructuring from omissions or hallucinations.\footnote{See Appendix~\ref{apdx:detailed_case_study} for case studies on distinguishing merging/splitting from omission/hallucination, and differentiating legitimate restructuring from those masking semantic loss.} Although splitting and merging are not categorized as pathological errors, effectively rectifying these structural nuances can yield substantial further performance gains.

As shown in our preliminary analysis in Table~\ref{tab:alignment_stats}, such errors are pervasive across model scales and persist even when sentence boundary constraints are explicitly introduced during generation~\cite{liu-etal-2025-improving,li-etal-2026-enhancing}. Notably, even strong proprietary models such as GPT-4o~\cite{openai-etal-gpt4o-2024} exhibit non-trivial rates of structural mismatch in Doc2Doc and Chunk2Chunk settings~\cite{liu-etal-2024-lost,shao-etal-2024-understanding}. These errors are particularly pronounced in long or complex documents. As input context chunk length scales up, the sentence alignment rate drops gradually. See Appendix~\ref{apdx:length_vs_star} for a detailed quantitative analysis.

This vulnerability is closely related to over-rejection \cite{xu-etal-2024-contrastive,xu-etal-2025-xalma} and reward hacking \cite{akter-etal-2026-detect}, where models adopt overly conservative strategies under safety or short-context biases. Although previous works~\cite{hu-etal-2025-source,domhan-zhu-2025-evaluation} attempt to investigate these length-related translation failures, their analyses are largely confined to superficial comparisons of input and output lengths. Such macroscopic perspectives fail to capture the fine-grained structural breakdowns. Importantly, structural misalignment is largely invisible to standard training objectives and evaluation metrics (e.g. COMET~\cite{rei-etal-2022-comet}), which primarily focus on semantic adequacy and fluency. As a result, Doc2Doc systems may achieve high metric scores while still omitting or hallucinating.

To address this limitation, we introduce \textbf{\underline{S}}entence \textbf{\underline{T}}ranslation \textbf{\underline{A}}lignment \textbf{\underline{R}}ate (\textbf{STAR}), an auxiliary metric that explicitly measures sentence-level structural fidelity in Doc2Doc translation. Building on STAR, we propose \textbf{\underline{STAR}}-masked \textbf{\underline{P}}reference \textbf{\underline{O}}ptimization (\textbf{StarPO}) framework that ranks document-level hypotheses by alignment quality and encourage structurally faithful generation. We further introduce a dynamic alignment masking strategy that downweights or excludes sentences that are already well aligned, allowing optimization to focus on misaligned segments such as omissions and hallucinations. 

In summary, our contributions are:
\begin{itemize}
\item We identify sentence-level structural misalignment as a key bottleneck in Doc2Doc translation and introduce \textbf{STAR} , a novel auxiliary metric for measuring structural fidelity.
\item We propose \textbf{\underline{STAR}}-masked \textbf{\underline{P}}reference \textbf{\underline{O}}ptimization (\textbf{StarPO}) with dynamic alignment masking to mitigate sentence omissions and hallucinations.
\item We demonstrate consistent and robust improvements in document-level translation quality across multiple domains and models.
\end{itemize}

\section{Methodology}

\begin{figure*}[t]
\centering

\includegraphics[width=1.0\textwidth]{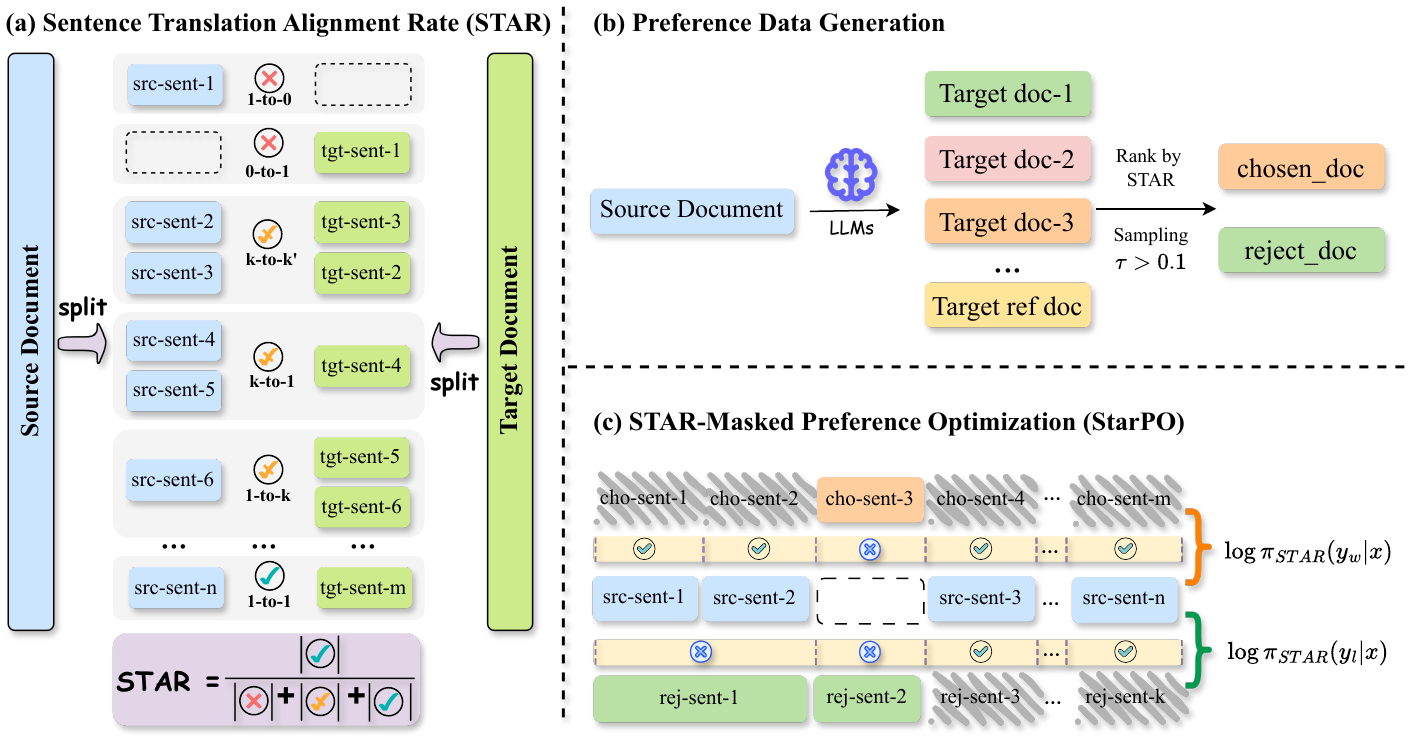}
\caption{Overview of the proposed framework. (a) Sentence Translation Alignment Rate (STAR): Source and target documents are segmented and aligned to compute STAR, defined as the ratio of strict 1-to-1 alignments. (b) Preference Data Generation: Translation candidates are sampled and ranked by their STAR scores. Pairs exhibiting a score disparity larger than $\tau$ are selected as chosen ($y_w$) and rejected ($y_l$) samples. (c) STAR-Masked Preference Optimization (StarPO): A sentence-level mask is applied to the CPO objective, excluding well-aligned (1-to-1) sentences to focus optimization exclusively on structurally problematic segments. }
\label{fig:main_pipeline}
\end{figure*}

In this section, we present a novel framework designed to enhance structural fidelity of document-level translation as illustrated in Figure~\ref{fig:main_pipeline}.

\subsection{Sentence Translation Alignment Rate (STAR)}
\label{sec:star}

We introduce Sentence Translation Alignment Rate (\textbf{STAR}), a metric for quantifying sentence-level structural fidelity in Doc2Doc translation.
As illustrate in Figure~\ref{fig:main_pipeline}(a), the computation of STAR proceeds in four steps:

\textbf{(1) Sentence Segmentation:} Source and target documents ($S$ and $T$) are segmented into sentences,  $S = \{s_1, \ldots, s_m\}$ and $T = \{t_1, \ldots, t_n\}$. Specifically, we segment sentences using SaT~\cite{frohmann-etal-2024-segment}.
 
\textbf{(2) Sentence-Level Alignment:} Sentences from the source and target are aligned to form disjoint alignment units $\mathcal{U} = \{u_1, \ldots, u_K\}$, where each unit $u_k = (\mathbf{s}_k, \mathbf{t}_k)$ may contain zero or more source sentences $\mathbf{s}_k \subseteq S$ and zero or more target sentences $\mathbf{t}_k \subseteq T$. These units are \textit{minimal} and cannot be further decomposed. Here, we use Bertalign~\cite{liu-zhu-2023-bertalign} for sentence-level alignment.
 
\textbf{(3) Unit Categorization:} Each alignment unit is classified based on the number of source and target sentences: (1) 1-to-1 ($\mathcal{U}_{1:1}$), where $|\mathbf{s}_k|=|\mathbf{t}_k|=1$; (2) Deletion ($\mathcal{U}_{1:0}$), where $|\mathbf{s}_k|\ge 1, |\mathbf{t}_k|=0$; (3) Insertion ($\mathcal{U}_{0:1}$), where $|\mathbf{s}_k|=0, |\mathbf{t}_k|\ge 1$; and (4) Complex ($\mathcal{U}_{\text{complex}}$), covering all other cases ($|\mathbf{s}_k| + |\mathbf{t}_k| > 2$ with $|\mathbf{s}_k|, |\mathbf{t}_k| \ge 1$).
 
\textbf{(4) STAR Computation:} STAR is the fraction of clean 1-to-1 units among all alignment units:
\begin{equation}
\small
\text{STAR}(S, T) = \frac{|\mathcal{U}_{1:1}|}{|\mathcal{U}_{1:1}| + |\mathcal{U}_{1:0}| + |\mathcal{U}_{0:1}| + |\mathcal{U}_{\text{complex}}|}.
\end{equation}

We also introduce a \textbf{relaxed} variant, $\text{STAR}_{\text{relax}}(S, T)$ , which treats $\mathcal{U}_{\text{complex}}$ as positive to exclusively penalize omissions and hallucinations while remaining tolerant of linguistically justified merging or splitting since complex alignments is also acceptable in real-world scenario:
\begin{equation}
\small
\text{STAR}_{\text{relax}}(S, T) = \frac{|\mathcal{U}_{1:1}| + |\mathcal{U}_{\text{complex}}| }{|\mathcal{U}_{1:1}| + |\mathcal{U}_{1:0}| + |\mathcal{U}_{0:1}| + |\mathcal{U}_{\text{complex}}|}.
\end{equation}

Thus, higher STAR indicates better sentence-level structural fidelity while lower STAR reflects deletions, insertions. Notably, STAR can also be computed directly via an LLM-as-a-judge approach, based on the four steps outlined above.\footnote{See Appendix~\ref{apdx:llm_judge_of_star} for detailed prompts.}

\subsection{Preference Data Generation }

To support STAR-masked preference optimization (StarPO), we construct a document-level preference dataset from automatically generated translation candidates. For each source document, we use GPT-4o~\cite{openai-etal-gpt4o-2024} to generate 5 translation candidates with a temperature of 1.0. Reference translation without sentence boundaries is also included in the candidate pool when available. 

For each souce document $S$, we compute the STAR score (Section~\ref{sec:star}) for all candidates and form a preference pair by selecting the candidate with the highest STAR score as the \emph{chosen} example ($T_w$) and the one with the lowest score as the \emph{rejected} example ($T_l$). To ensure meaningful supervision, a pair $(T_w, T_l)$ is retained only if the STAR score difference exceeds a threshold $\tau$, i.e., $|\text{STAR}(S, T_w) - \text{STAR}(S, T_l)| > \tau$. In our experiments, we set $\tau = 0.1$.\footnote{For more discussion of $\tau$, see Appendix ~\ref{apdx:tau}. }

\subsection{STAR-Masked Preference Optimization (StarPO)}

Following recent studies~\cite{wang-etal-2026-beyond,agrawal-etal-2024-modeling,xu-etal-2024-contrastive,sun-etal-2025-enhancing-machine}, we adopt a two-stage training paradigm. We first perform supervised fine-tuning (SFT) on high-quality parallel corpora to establish a warm-started policy $\pi_{\text{SFT}}$. We then apply preference optimization to further align the model with structural constraints.

\paragraph{Background: Contrastive Preference Optimization.} 
Following contrastive preference optimization (CPO)~\cite{xu-etal-2024-contrastive,xu-etal-2025-xalma}, we assume a preference dataset $\mathcal{D}=\{(S, T_w, T_l)\}$, where $S$ is a source document and $(T_w, T_l)$ denote a preferred and dis-preferred translation, respectively. CPO optimizes the policy $\pi_\theta$ by maximizing the likelihood margin between the preferred and dis-preferred candidates. The standard objective is formulated as:
\begin{equation}
\small
\begin{split}
\mathcal{L}_{\text{CPO}}(\pi_\theta) = & -\mathbb{E}_{\mathcal{D}} \Big[ \log \sigma \big( \beta \log \pi_\theta(T_w|S) \\
& \quad - \beta \log \pi_\theta(T_l|S) \big) \Big] \\
& - \mathbb{E}_{\mathcal{D}} \big[ \log \pi_\theta(T_w|S) \big],
\end{split}
\end{equation}
where $\sigma$ is sigmoid function and $\beta$ is a hyperparameter controlling strength of preference signal. Typically, $\log \pi_\theta(T|S)$ is computed as the sum of log-probabilities over all tokens in document $T$.

\paragraph{STAR-Masked Objective.} 
Directly applying CPO at the document-level treats all tokens equally, regardless of whether they correspond to structurally correct translations or not. To focus learning on sentence-level structural issues (e.g., hallucinations, omissions, or complex alignments), we introduce a STAR-based masking strategy. 

Given a target document $T$ which is segmented into $n$ sentences $T = \{t_1, \ldots, t_n\}$ and the alignment units defined in Section~\ref{sec:star}, we define a sentence-level mask $\mathcal{M}(t_j)$ that indicates whether sentence $t_j$ should contribute to preference loss: 


\begin{equation}
\small
\mathcal{M}(t_j) = 1 - \mathbb{I}_{1:1}(t_j),\end{equation}
where $\mathbb{I}_{1{:}1}(t_j)=1$ if sentence $t_j$ belongs to a clean 1-to-1 alignment unit and $0$ otherwise. 

As a result, well-aligned sentences receive a mask value of 0, whereas sentences associated with structural mismatches receive a value of 1.

Using this mask, we define STAR-Masked log-likelihood, $\log \pi_{\text{STAR}}(T|S)$, which aggregates token-level probabilities strictly for non-1-to-1 sentences:
\begin{equation}
\small
\begin{split}
    \log \pi_{\text{STAR}}(T|S) &= \sum_{j=1}^n \bigg[ \mathcal{M}(t_j) \cdot \\
    &\quad \sum_{k=1}^{|t_j|} \log \pi_\theta\left(t_{j,k} | t_{<j}, t_{j,<k}, S\right) \bigg],
\end{split}
\end{equation}
where $|t_j|$ is the number of tokens in sentence $t_j$, $t_{j,k}$ denotes the $k$-th token in the $t_j$, $t_{<j}$ represents all target sentences preceding $t_j$, and $t_{j,<k}$ denotes the generated tokens in $t_j$, respectively. The inner summation corresponds to the standard token-level log-likelihood for sentence $t_j$.

We then replace the standard document-level likelihood in CPO with the masked likelihood to obtain the STAR-masked preference loss:
\begin{equation}
\small
\begin{split}
\mathcal{L}_{\text{StarPO}}(\pi_\theta) = & -\mathbb{E}_{\mathcal{D}} \Big[ \log \sigma \big( \beta \log \pi_{\text{STAR}}(T_w|S) \\
& \quad - \beta \log \pi_{\text{STAR}}(T_l|S) \big) \Big] \\
& - \mathbb{E}_{\mathcal{D}} \big[ \log \pi_{\text{STAR}}(T_w|S) \big].
\end{split}
\end{equation}

\section{Experimentation}
\subsection{Experimental Settings}

\paragraph{Datasets.} Following recent works~\cite{wang-etal-2025-delta,liu-etal-2025-improving,cui-etal-2024-efficiently}, we evaluate on both news and web novels from WMT. Specifically, we use News-Commentary v18.1 from WMT25 for English (En) $\Leftrightarrow$ \{ Chinese (Zh), German (De), Russion (Ru), Spanish (Es)\} and Chinese (Zh) $\Leftrightarrow$ German (De) and Guofeng ~\cite{wang-etal-2024-benchmarking}\footnote{\href{https://github.com/longyuewangdcu/GuoFeng-Webnovel/}{\nolinkurl{github.com/longyuewangdcu/GuoFeng-Webnovel/}}} from WMT25\footnote{\href{https://www2.statmt.org/wmt25/translation-task.html}{\nolinkurl{www2.statmt.org/wmt25/translation-task.html}}} for Chinese (Zh) $\Leftrightarrow$ \{English (En), German (De), Russian (Ru)\} translation. We use the training sets to fine-tune the LLMs and to construct preference pairs. Detailed dataset statistics are provided in Appendix~\ref{apdx:dataset stat}.

\paragraph{Models and Implementation Details. } We use three open-source instruction-tuned large language models: LLaMA-3.1-8B-Instruct~\cite{team-etal-2024-llama3}\footnote{\href{https://huggingface.co/meta-llama/Llama-3.1-8B-Instruct}{\nolinkurl{huggingface.co/meta-llama/Llama-3.1-8B-Instruct}}}, Qwen-2.5-7B-Instruct~\cite{qwen2.5}\footnote{\href{https://huggingface.co/Qwen/Qwen2.5-7B-Instruct}{\nolinkurl{huggingface.co/Qwen/Qwen2.5-7B-Instruct}}} and Qwen-3-4B-Instruct\footnote{\href{https://huggingface.co/Qwen/Qwen3-4B-Instruct-2507}{\nolinkurl{huggingface.co/Qwen/Qwen3-4B-Instruct-2507}}}. See Appendix~\ref{apdx:implementation_details} for more implementation details.

\paragraph{Systems.} For comprehensive comparison, we evaluate two categories of systems:

{(1) Training Paradigms.} We report the performance of the original instruct versions of the LLMs without further fine-tuning (referred to as \textbf{Base}), as well as models further fine-tuned with supervised fine-tuning (\textbf{+SFT}) following \citet{li-etal-2026-enhancing}. We use our full training set, further optimized with standard contrastive preference optimization (\textbf{+CPO}), and our proposed STAR-masked preference optimization (\textbf{+StarPO}).

{(2) State-of-the-Art and Competitive Systems.} We additionally compare against strong document-level translation systems, including Tower-plus-9B~\cite{rei-etal-2024-tower}, GPT-4o~\cite{openai-etal-gpt4o-2024}, and Deepseek-R1~\cite{deepseek-2025-deepseek}.

\begin{table*}[t]
\centering
\small
\resizebox{\textwidth}{!}{
\begin{NiceTabular}{lccccccccccc}
\toprule
\Block[l]{2-1}{\textbf{System}} & \Block[c]{1-2}{\textbf{Zh $\Leftrightarrow$ En}} & & \Block[c]{1-2}{\textbf{De $\Leftrightarrow$ En}} & & \Block[c]{1-2}{\textbf{De $\Leftrightarrow$ Zh}} & & \Block[c]{1-2}{\textbf{Ru $\Leftrightarrow$ En}} & & \Block[c]{1-2}{\textbf{En $\Leftrightarrow$ Es}} & & \Block[l]{2-1}{\textbf{Avg.}}\\
\cmidrule(lr){2-3}\cmidrule(lr){4-5}\cmidrule(lr){6-7}\cmidrule(lr){8-9}\cmidrule(lr){10-11}
& $\Rightarrow$ & $\Leftarrow$ & $\Rightarrow$ & $\Leftarrow$ & $\Rightarrow$ & $\Leftarrow$& $\Rightarrow$ & $\Leftarrow$& $\Rightarrow$ & $\Leftarrow$ \\ 
\midrule

\rowcolor[gray]{0.9}
\Block[c]{1-12}{\textsc{Llama-3.1-8B-Instruct}} \\ 
Base & 72.23 & 74.89 & 71.60 & 80.13 & 72.49 & 58.80 & 80.65 & 80.75 & 80.53 & 78.62 & 75.07 \\
+ SFT & 76.81 & 80.01 & 83.79 & 81.47 & 75.42 & 75.41 & 82.37 & 80.69 & 80.66 & 83.49 & 80.01 \\
\hspace{1em}+ CPO        & 81.10 & 79.94 & 83.98 & 82.42 & 75.50 & 75.05 & 82.51 & 81.43 & 81.45 & 84.60 & 80.80\\
\hspace{1em}+ StarPO   & \colorbox{blue!15}{81.55} & \colorbox{blue!15}{80.11} & \colorbox{blue!15}{\underline{84.05}} & \colorbox{blue!15}{82.50} & \colorbox{blue!15}{76.33} & \colorbox{blue!15}{\textbf{77.08}} & \colorbox{blue!15}{82.53} & \colorbox{blue!15}{81.71} & \colorbox{blue!15}{82.03} & \colorbox{blue!15}{\underline{84.91}}  & \colorbox{blue!15}{\underline{81.28}}  \\ 
\midrule
\rowcolor[gray]{0.9}
\Block[c]{1-12}{\textsc{Qwen2.5-7B-Instruct}} \\ 
Base        & 81.57 & 80.40 & 83.31 & 79.09 & 77.60 & 73.97 & 81.94 & 78.35 & 81.33 & 84.34 & 80.19 \\
+ SFT        & \underline{82.01} & 80.89 & 83.61 & 80.02 & 77.68 & 74.45 & 82.26 & 81.65 & 81.75 & 84.73 & 80.91  \\
\hspace{1em}+ CPO        & 81.94 & \underline{80.97} & 84.01 & 81.14 & \textbf{78.27} & 74.56 & \underline{82.59} & 81.79 & 81.65 & 84.71 & 81.16\\

\hspace{1em}+ StarPO   & \textbf{82.27} & \textbf{\colorbox{blue!15}{81.33}} & \textbf{\colorbox{blue!15}{84.06}} & \colorbox{blue!15}{81.89} & \underline{78.22} & 74.58 & \textbf{\colorbox{blue!15}{82.70}} & \colorbox{blue!15}{82.12} & 81.79 & \textbf{\colorbox{blue!15}{85.21}} & \bf \colorbox{blue!15}{81.42} \\ 
\midrule
\rowcolor[gray]{0.9}
\Block[c]{1-12}{\textsc{Qwen3-4B-Instruct}} \\ 
Base        & 81.47 & 80.31 & 83.63 & 80.87 & 77.43 & 76.78 & 81.58 & 84.03 & 80.21 & 84.01 & 81.03 \\
+ SFT        & 81.71 & 80.88 & 83.65 & 81.50 & 77.37 & 76.87 & 82.19 & 84.04 & 80.46 & 84.70 & 81.34 \\
\hspace{1em}+ CPO        & 81.77 & 80.90 & 83.56 & 81.55 & 77.79 & 76.92 & 82.23 & 84.01 & 81.20 & 84.67 & \underline{81.46} \\
\hspace{1em}+ StarPO & \colorbox{blue!15}{\underline{82.24}} & \colorbox{blue!15}{\underline{81.17}} & 83.84 & \colorbox{blue!15}{82.07} & 78.00 & \underline{76.93} & \colorbox{blue!15}{82.43} & 84.21 & \colorbox{blue!15}{81.48} & \colorbox{blue!15}{\underline{84.93}} & \textbf{81.73} \\ 
\midrule
\rowcolor[gray]{0.9}
\Block[c]{1-12}{\textsc{Other Systems}} \\ 
Tower+ & 80.53 & 80.18 & 84.01 & 82.57 & 75.85 & \underline{76.34} & 82.58 & 81.66 & 82.46 & 84.81 & 81.10 \\
GPT-4o      & 77.42 & 80.64 & 83.87 & \underline{82.84} & 77.88 & 69.32 & 82.36 & \underline{84.73} & \textbf{83.31} & 84.83 & 80.72 \\
Deepseek-R1 & 79.52 & 79.10 & 82.19 & \textbf{82.90} & 77.92 & 75.96 & 81.90 & \textbf{84.97} & \underline{82.56} & 82.71 & 80.97 \\
\bottomrule
\end{NiceTabular}
}
\caption{Performance in dCOMET score on the News-Commentary test set. \textbf{Bold} scores represent the global best performance and \underline{underlined} scores represent the global second-best performance. \colorbox{blue!15}{Blue text background} indicates that the improvement over the origin Base model achieves at least 85\% accuracy with the human judgment~\cite{xu-etal-2024-contrastive,kocmi-etal-2024-navigating}. Specifically, the improvement needs a minimin of $\ge 0.71$ for \texttt{wmt22-comet-da}. }
\label{table:main_comet_results}
\end{table*}

\begin{table*}[t]
\centering
\small
\resizebox{\textwidth}{!}{
\begin{NiceTabular}{lccccccccccc}
\toprule
\Block[l]{2-1}{\textbf{System}} & \Block[c]{1-2}{\textbf{Zh $\Leftrightarrow$ En}} & & \Block[c]{1-2}{\textbf{De $\Leftrightarrow$ En}} & & \Block[c]{1-2}{\textbf{De $\Leftrightarrow$ Zh}} & & \Block[c]{1-2}{\textbf{Ru $\Leftrightarrow$ En}} & & \Block[c]{1-2}{\textbf{En $\Leftrightarrow$ Es}} & & \Block[l]{2-1}{\textbf{Avg.}}\\
\cmidrule(lr){2-3}\cmidrule(lr){4-5}\cmidrule(lr){6-7}\cmidrule(lr){8-9}\cmidrule(lr){10-11}
& $\Rightarrow$ & $\Leftarrow$ & $\Rightarrow$ & $\Leftarrow$ & $\Rightarrow$ & $\Leftarrow$& $\Rightarrow$ & $\Leftarrow$& $\Rightarrow$ & $\Leftarrow$ \\ 
\midrule

\rowcolor[gray]{0.9}
\Block[c]{1-12}{\textsc{Llama-3.1-8B-Instruct}} \\ 
Base            & 19.25 & 20.03 & 36.37 & 27.08 & 22.89 & 11.56 & 40.31 & 22.21 & 42.15 & 40.25 & 28.21 \\
+ SFT           & 21.55 & 39.36 & 44.20 & 28.54 & 30.62 & 15.43 & 41.65 & 23.78 & 42.60 & 43.49 & 33.12 \\
\hspace{1em}+ CPO         & 30.98 & 38.77 & 44.02 & 28.27 & 34.98 & \underline{16.69} & \underline{42.11} & 24.45 & 43.14 & 43.52 & 34.69 \\
\hspace{1em}+ StarPO      & \colorbox{blue!15}{32.88} & \colorbox{blue!15}{40.20} & \bf \colorbox{blue!15}{44.36} & 28.77 & \colorbox{blue!15}{35.90} & \bf \colorbox{blue!15}{17.94} & \textbf{42.45} & \colorbox{blue!15}{25.72} & 43.99 & \colorbox{blue!15}{\underline{43.68}} & \colorbox{blue!15}{\bf 35.59} \\ 
\midrule
\rowcolor[gray]{0.9}
\Block[c]{1-12}{\textsc{Qwen2.5-7B-Instruct}} \\ 
Base            & 30.14 & 41.63 & 41.19 & 24.28 & 36.29 & 13.29 & 40.65 & 21.25 & 41.28 & 43.03 & 33.30 \\
+ SFT           & 29.62 & 42.27 & 42.97 & 25.34 & 35.08 & 14.45 & 41.63 & 24.05 & 41.57 & 43.59 & 34.06 \\
\hspace{1em}+ CPO         & 30.90 & \underline{43.34} & 43.71 & 25.78 & \textbf{37.46} & 14.52 & 41.87 & 24.10 & 41.71 & 43.13 & 34.65 \\
\hspace{1em}+ StarPO      & \bf \colorbox{blue!15}{35.23} & \bf 44.12 & \underline{44.34} & 26.52 & \underline{36.34} & 14.92 & 42.05 & \colorbox{blue!15}{25.04} & 42.14 & \bf 44.02 & \underline{35.47} \\ 
\midrule
\rowcolor[gray]{0.9}
\Block[c]{1-12}{\textsc{Qwen3-4B-Instruct}} \\ 
Base            & 27.71 & 41.55 & 42.22 & 26.01 & 31.68 & 14.27 & 38.34 & 25.00 & 40.58 & 43.91 & 33.13 \\
+ SFT           & 30.35 & 42.07 & 43.61 & 27.09 & 33.72 & 14.79 & 41.41 & 27.37 & 42.79 & 44.09 & 34.73 \\
\hspace{1em}+ CPO         & 33.68 & 42.54 & 43.80 & 27.24 & 34.41 & 15.33 & 41.80 & 28.11 & 42.68 & 44.46 & 35.41 \\
\hspace{1em}+ StarPO      & 35.93 & 43.42 & 43.41 & 28.11 & 33.66 & 15.61 & 42.09 & 28.12 & 42.62 & 45.14 & 35.81 \\ 
\midrule
\rowcolor[gray]{0.9}
\Block[c]{1-12}{\textsc{Other Systems}} \\ 
Tower+            & \underline{33.37} & 41.89 & 43.16 & 28.52 & 34.22 & 15.08 & 41.97 & 26.95 & 42.85 & 43.16 & 35.12 \\
GPT-4o            & 27.70 & 42.06 & 42.83 & \underline{29.62} & 35.38 & 12.70 & 41.28 & \bf 28.55 & \bf 44.24 & 43.12 & 34.75 \\
Deepseek-R1       & 32.33 & 39.27 & 41.68 & \bf 29.83 & 33.59 & 14.00 & 40.44 & \underline{27.28} & \underline{44.06} & 41.64 & 34.41 \\
\bottomrule
\end{NiceTabular}
}
\caption{Performance in d-BLEU score on the News-Commentary test set. Unlike dCOMET, d-BLEU evaluates each document as a single continuous string without sentence alignment. \colorbox{blue!15}{Blue text background} denotes a significant improvement over the Base model ($\ge 3.35$ BLEU).}
\label{table:main_bleu_results}
\end{table*}
\begin{table}[t]
\centering
\small
\resizebox{\columnwidth}{!}{
\begin{NiceTabular}{lcccccc}
\toprule
\Block[l]{2-1}{\textbf{System}} & \Block[c]{1-2}{\textbf{Zh $\Leftrightarrow$ En}} & & \Block[c]{1-2}{\textbf{Zh $\Leftrightarrow$ De}} & & \Block[c]{1-2}{\textbf{Zh $\Leftrightarrow$ Ru}}\\
\cmidrule(lr){2-3}\cmidrule(lr){4-5}\cmidrule(lr){6-7}
& $\Rightarrow$ & $\Leftarrow$ & $\Rightarrow$ & $\Leftarrow$ & $\Rightarrow$ & $\Leftarrow$ \\ 
\midrule
\rowcolor[gray]{0.9}
\Block[c]{1-7}{\textsc{Llama-3.1-8B-Instruct}}                                                   \\ 
Base & 53.66 & 64.33 & 52.31 & 66.36 & 60.28 & 62.04 \\
+ SFT & 62.70 &  65.42 & 67.96 & 73.33 & 75.27 & 72.75  \\
\hspace{0.5em}+ CPO  & 63.01 & 68.70 & 69.04 & 73.95 & 75.22 & 72.92 \\
\hspace{0.5em}+ StarPO & \colorbox{blue!15}{63.43} & \colorbox{blue!15}{72.17} & \bf \colorbox{blue!15}{72.15} & \colorbox{blue!15}{74.61} & \bf \colorbox{blue!15}{77.50} & \colorbox{blue!15}{73.79} \\ \midrule
\rowcolor[gray]{0.9}
\Block[c]{1-7}{\textsc{Qwen2.5-7B-Instruct}}                                                  \\ 
Base & 70.22 & 71.99 & 64.11 & 73.04 &  55.74 &  73.31   \\
+SFT & \underline{70.42} & 69.54 & 66.66 &  73.03 & 75.86  & 72.23   \\
\hspace{0.5em}+CPO & 70.20 &  71.23  & 66.51 &  73.69 & 74.63 & 72.78  \\
\hspace{0.5em}+StarPO & \bf 70.64 &  72.13 & \colorbox{blue!15}{\underline{70.72}} & \colorbox{blue!15}{73.94} & \colorbox{blue!15}{\underline{77.46}} &  \colorbox{blue!15}{75.00} \\ \midrule
\rowcolor[gray]{0.9}
\Block[c]{1-7}{\textsc{Qwen3-4B-Instruct}}                                                  \\ 
Base & 63.61 & 68.64 & 68.01& 73.76 & 72.98 & 73.04 \\
+SFT & 68.61 & 69.77 & 69.41 & \underline{75.08} & 76.48 & \underline{75.06} \\
\hspace{0.5em}+CPO & 68.77 & 71.02 & 69.01 & 74.84 & 76.49 & 73.83 \\
\hspace{0.5em}+StarPO & \colorbox{blue!15}{70.13} & \colorbox{blue!15}{\underline{72.21}} & \colorbox{blue!15}{70.20} & \bf \colorbox{blue!15}{75.20} & \colorbox{blue!15}{76.53} & \bf \colorbox{blue!15}{75.44} \\ \midrule
\rowcolor[gray]{0.9}
\Block[c]{1-7}{\textsc{Other Systems}}  \\ 
Tower+ & 69.39 & 65.92 & 54.41 & 72.89 & 70.39  & 66.48    \\
GPT-4o    & 69.58 & \bf 73.61 & 53.24 & 73.91 &  55.35 & 70.65 \\ 
Deepseek & 62.86 & 69.58 & 64.51 & 70.30 & 71.53 & 67.35\\
\bottomrule
\end{NiceTabular}
}
\caption{Performance in dCOMET on Guofeng test set.}
\label{table:COMET on Guofeng testset}
\end{table}

\paragraph{Metrics.} We evaluate translation quality using document-level COMET (dCOMET)~\cite{vernikos-etal-2022-embarrassingly}, computed by \texttt{wmt22-comet-da}~\cite{rei-etal-2022-comet}. Specifically, since document-level evaluation requires sentence-level alignments~\cite{vernikos-etal-2022-embarrassingly}, we apply the alignment strategy described in Section \ref{sec:star}. Following \citet{zouhar-etal-2024-pitfalls} and \citet{zebaze-etal-2025-context}, using our settings in STAR$_{\text{relax}}$, we assign a score of 0 COMET to unaligned segments, including omissions (1-to-0) and hallucinations (0-to-1), while complex mappings (1-to-$k$, $k$-to-1, $k$-to-$k'$) \textbf{retain} their computed COMET scores. To provide a more holistic evaluation, we use d-BLEU, which bypasses sentence-splitting tools by treating the entire document as a single continuous string for both input and output. This ensures that the evaluation remains uninfluenced by potential sentence-boundary artifacts. See Appendix~\ref{apdx:Detailed_metric_scores} for our proposed STAR and STAR$_{\text{relax}}$ scores.

\subsection{Experimental Results}

Table \ref{table:main_comet_results} and Table~\ref{table:main_bleu_results} present the results on News-Commentary dataset. Across all backbones, supervised fine-tuning (+SFT) yields consistent and substantial improvement over the base models on most language pairs. Building on this, standard CPO (+CPO) further enhances translation quality. Our proposed StarPO consistently outperforms standard CPO and achieves the best average performance across models. For instance, on LLaMA-3.1, StarPO obtains an average improvement of 0.48 COMET over standard CPO. Importantly, these gains are stable across model scales. With StarPO, the compact model outperforms strong SOTA and competitive systems, including Tower-plus-9B, GPT-4o, and DeepSeek-R1. This highlights the effectiveness of structural alignment preference optimization for document-level translation. 

Table~\ref{table:COMET on Guofeng testset} present dCOMET scores on Guofeng dataset.\footnote{See Appendix~\ref{apdx:guofeng_bleu} for detailed BLEU scores on Guofeng datasets.} It is worth noting that web novel translation typically does not adhere to rigid 1-to-1 literalism, often employing complex sentence mappings for stylistic flow. Although StarPO enforces 1-to-1 constraints during training, our evaluation remains inclusive of valid complex mappings.  StarPO’s training constraints prevent pathologies like language mismatches without stifling this flexibility, as evidenced by restoring coherence where LLaMA-3.1-base fails on Zh$\Rightarrow$De (52.31 COMET). These results indicate that smaller models, when trained with structural-alignment preference optimization, can outperform massive generalist models on complex, stylized document-level translation tasks.

\subsection{Comparison with Alternative Optimization Baselines}

\begin{table}[t]
\centering
\small
\begin{NiceTabular}{lcc}
\toprule
\textbf{Method} & \textbf{Zh $\Rightarrow$ En} & \textbf{En $\Rightarrow$ Zh} \\
\midrule
\rowcolor{gray!15} 
\Block[l]{1-3}{\textit{Data Ranking Metrics}} \\
BLEU Ranking & 76.55 & 79.71 \\
COMET Ranking & 81.01 & 79.78 \\
COMETKiwi Ranking & 80.56 & 79.73\\
Word-level Coverage & 75.94 & 79.46 \\
\midrule
\rowcolor{gray!15} 
\Block[l]{1-3}{\textit{Online RL Strategies}} \\
GRPO (\textit{w.} COMET) & 77.73 & 79.50\\
GRPO (\textit{w.} STAR) & 77.79 & 79.44\\
GRPO (\textit{w.} BLEU) & 77.81 & 79.52 \\
GRPO (\textit{w.} COMET + STAR) & 77.76 & 79.76 \\
GSPO (\textit{w.} COMET) & 77.94 & 79.77 \\
GSPO (\textit{w.} STAR) & 80.16 & 80.00 \\
GSPO (\textit{w.} BLEU) & 77.96 & 79.81 \\
GSPO (\textit{w.} COMET + STAR) & 77.72 & 79.97 \\
\midrule
\rowcolor{gray!15} 
\Block[l]{1-3}{\textit{Our Variants \& Ablation Studies }} \\
SFT on preference data & 80.41 & 80.02 \\
STAR (Relaxed) & 80.09 & 79.59 \\
Random Mask (Sentence level) & \underline{81.18} & 80.05 \\
Random Mask (Token level) & 81.17 & \underline{80.06}\\
CPO & 81.10 & 79.94 \\
\textbf{Ours (StarPO)} & \bf 81.55 & \bf 80.11 \\
\bottomrule
\end{NiceTabular}
\caption{COMET scores comparison against alternative optimization strategies. Word-level Coverage is calculated via WSPAlign \cite{wu-etal-2023-wspalign,wu-etal-2024-word} at the aligned sentence-pair level.}
\label{table:comprehensive_comparison}
\end{table}

To further illustrate the effectiveness of StarPO, we compare it against a diverse set of alternative optimization baselines. We use LLaMA-3.1-8B-Instruct on Zh $\Leftrightarrow$ En as a representative setting and report the results in Table~\ref{table:comprehensive_comparison}.

\paragraph{Effect of Ranking Metrics on Preference Data Construction.}
We evaluate the impact of ranking signals by substituting STAR with COMET, COMETKiwi, BLEU, and word-level alignment coverage~\cite{wu-etal-2024-word,wu-etal-2023-wspalign}, under a strictly controlled data budget identical to our main experiment. As shown in Table~\ref{table:comprehensive_comparison}, STAR significantly outperforms all baselines despite the equalized data size. Notably, its superiority over sentence-level methods~\citep{he-etal-2024-improving,agrawal-etal-2024-modeling,tang-etal-2025-mitigating} confirms that discourse-level structural fidelity is best captured via sentence-to-sentence correspondence.

\paragraph{Comparison with RL.}
We further benchmark our method against standard online RL strategies (GRPO~\cite{shao-etal-2024-deepseekmath}, GSPO~\cite{zheng-etal-2025-gspo}). Following recent approaches~\cite{feng-etal-2025-mt-r1,feng-etal-2026-mt3,he-etal-2025-r1t1}, we directly inject COMET, STAR and BLEU scores into the reward function for these baselines. As observed in the middle section of Table~\ref{table:comprehensive_comparison}, while online methods utilizing self-generated data generally underperform our offline framework, GSPO paired with STAR achieves competitive results (80.16 COMET). This indicates that while our offline optimization (StarPO) is more effective, STAR nevertheless functions as a robust and high-quality reward signal within RL paradigms. Detailed analyses and comparisons with other standard offline preference optimization algorithms (e.g., DPO~\cite{rafailov-etal-2023-dpo}, SimPO~\cite{meng-etal-2024-simpo}) are provided in Appendix~\ref{apdx:offline_po}.

\paragraph{Ablation Studies.} To further dissect the source of our gains, we examine several specific variants in the bottom section of Table~\ref{table:comprehensive_comparison}: 

\textbf{(1) SFT vs. Preference Optimization:} Training solely on preferred responses ($y_w$) yields strong performance, yet our full framework consistently outperforms this SFT baseline , proving that the contrastive signal in preference optimization is more effective than simple supervised imitation.

\textbf{(2) Strict vs. Relaxed STAR:} Using the relaxed STAR metric degrades performance, as it dilutes the discriminative power needed to establish a sufficient score margin for data selection during training (see Appendix~\ref{apdx:star_distribution} for score distribution analysis). 

\textbf{(3) Alignment-aware vs. Random Masking:} We further investigate the effect of loss masking by randomly masking out 90\% of tokens\footnote{See Appendix~\ref{apdx:mask_distribution} for the distribution of unmasked token ratios during training.} at both the sentence and token levels. While random token- or sentence-level masking acts as a helpful regularizer~\cite{gu-etal-2025-maskdpo}, StarPO’s semantic-aware masking consistently achieves superior results. This demonstrates that focusing the optimization objective on structurally misaligned segments provides a more effective learning signal than random selection.\footnote{See Appendix~\ref{apdx:masking_analysis} for theoretical analysis.}

\section{Discussions}

\subsection{Sensitivity to Structural Pathologies}

\begin{table}[t]
\centering
\small
\setlength{\tabcolsep}{3pt}
\begin{NiceTabular}{lcc}
\toprule
\Block{2-1}{\bfseries Methods}
& \Block{1-2}{\bfseries Correlation ($\rho, \uparrow$)} \\

\cmidrule(lr){2-3} 
& \bfseries Strict $1$-to-$1$
& \bfseries Relaxed $N$-to-$M$ \\
\midrule

\rowcolor{gray!15}
\Block[l]{1-3}{\textit{Existing Alignment Methods}} \\
Align-then-Slide & 0.3804 & 0.5168 \\
SEGALE & 0.4193 & 0.4760 \\

\midrule
\rowcolor{gray!15}
\Block[l]{1-3}{\textit{Simple Metrics}} \\
Token Count Ratio & 0.0218 & 0.0211 \\ 
Sentence Count Ratio & 0.1218 & 0.1780 \\
Sentence Count Difference & 0.0649 & 0.1647 \\

\midrule
\rowcolor{gray!15}
\Block[l]{1-3}{\textit{Ours (STAR Variants)}} \\
\textbf{STAR (Section \ref{sec:star})} & \bf 0.5808 & \bf 0.5774 \\
\quad \textit{w/o} SaT (use Spacy) & \underline{0.5644} & \underline{0.5663} \\
\quad \textit{w/o} LaBSE (use M3) & 0.5104 & 0.5479 \\

\bottomrule
\end{NiceTabular}
\caption{
Spearman correlation ($\rho$) between LLM-annotated alignment quality and various evaluation methods (including STAR variants, Align-then-Slide~\cite{guo-etal-2025-align}, SEGALE~\cite{wang-etal-2025-extending}, and length-based metrics) under both strict $1$-to-$1$ and relaxed $N$-to-$M$ matching criteria.
}
\label{table:correlation_analysis}
\end{table}

To verify STAR’s capability in identifying structural pathologies, we evaluate its performance. Starting from WMT24~\cite{kocmi-etal-2024-preliminary} test set,\footnote{\href{https://github.com/wmt-conference/wmt24-news-systems}{\nolinkurl{github.com/wmt-conference/wmt24-news-systems}}} we introduce several LLMs including GPT-4o, Gemini-1.5-Flash, Hunyuan\cite{sun-etal-2024-hunyuan}, Qwen3-235B\footnote{\href{https://huggingface.co/Qwen/Qwen3-235B-A22B}{\nolinkurl{huggingface.co/Qwen/Qwen3-235B-A22B}}}, and DeepSeek-R1 to produce translations across several language pairs: En $\Leftrightarrow\{$De, Zh, Ru, Es$\}$. To establish a reliable ground truth, we employ Gemini-2.5-Pro to annotate the structural alignment score (i.e. LLM implementation of STAR). We then compute Spearman ($\rho$) between the metric scores and the labels in two settings: strict mode and relaxed mode. Specifically, the strict mode treats only 1-to-1 mappings as valid alignments, enforcing a rigid sentence-level correspondence. In contrast, the relaxed mode accounts for "reasonable restructuring" by accepting linguistically justified $N$-to-$M$ mappings (e.g., merging or splitting) as valid, thereby penalizing only the pathological 1-to-0 omissions and 0-to-1 hallucinations. The results are shown in Table \ref{table:correlation_analysis}.  

\paragraph{Comparison with Existing Alignment Methods.} 

Both Align-then-Slide~\cite{guo-etal-2025-align} and SEGALE~\cite{wang-etal-2025-extending} utilize sentence alignment between source and target text for document-level evaluation. We compute STAR scores using their intermediate alignment results for a fair comparison. As shown in Table \ref{table:correlation_analysis}, our implementation achieves superior correlation with structural noise. This highlights that precise, fine-grained sentence alignment is a critical prerequisite for reliable document-level structural assessment.

\paragraph{Comparison with length-based Metrics.} We compare STAR with length-based metrics in prior works for detecting omissions, hallucinations or collapse~\cite{peng-etal-2025-investigating,guerreiro-etal-2023-looking,shao-etal-2024-understanding,hu-etal-2025-source,domhan-zhu-2025-evaluation}, including \textit{Token Count Ratio}, \textit{Sentence Count Ratio}, and \textit{Sentence Count Difference}. As shown in Table~\ref{table:correlation_analysis}, these metrics exhibit weak correlations ($\rho < 0.2$), failing to detect structural noise when overall document length is preserved. 

\paragraph{Ablation Study on STAR Components.} We verify the robustness of STAR by swapping its components. Replacing SaT with Spacy\footnote{\url{https://github.com/explosion/spaCy}} or switching the LaBSE encoder to M3~\cite{chen-etal-2024-m3} results only marginal performance fluctuations. These results demonstrate that STAR's effectiveness is relatively robust.

\subsection{Analysis of Structural Alignment and Semantic Accuracy}

While complex alignments ($k$-to-$k'$, including merging and splitting) are sometimes feasible, they frequently mask local semantic distortions or partial omissions. To investigate this, we conduct a fine-grained COMET analysis on Zh $\Rightarrow$ En, categorizing source sentences into three groups: \textbf{Pathological Errors} (1-to-0/0-to-1), \textbf{Complex Restructuring} (merging/splitting/swapping), and \textbf{Preserved} 1-to-1 mappings. 

Table~\ref{table:fine_grained_comet} shows that StarPO’s improvement stems from mitigating pathological errors and, more significantly, rectifying \textit{Complex} restructuring. While SFT offers marginal gains, StarPO substantially boosts the COMET score for complex segments (71.87 $\rightarrow$ 81.52), confirming that enforcing structural 1-to-1 correspondence recovers semantic details typically lost in merging or splitting. 

\begin{table}[ht]
\centering
\small
\setlength{\tabcolsep}{1pt}
\begin{NiceTabular}{lccccccc}
\toprule
\multirow{2}{*}{\bf Pattern} & \multicolumn{2}{c}{\bf Pathological} & \multicolumn{2}{c}{\bf Complex} & \multicolumn{2}{c}{\bf 1-to-1} & \bf Overall \\
\cmidrule(lr){2-3} \cmidrule(lr){4-5} \cmidrule(lr){6-7}
 & \bf \% & \bf COMET &\bf  \% & \bf COMET & \bf \% & \bf COMET & \bf COMET \\
 \midrule
Base & 3.25 & 0.00 & 4.16 & 71.87 & 92.59 & 75.78 & 73.15 \\
SFT & 3.14 & 0.00 & 3.13 & 74.81 & 93.73 & 79.45 & 76.81 \\
StarPO & 2.09 & 0.00 & 2.12 & 81.52 & 95.79 & 83.33 & 81.55 \\
\bottomrule
\end{NiceTabular}
\caption{Fine-grained analysis on Zh $\Rightarrow$ En. }
\label{table:fine_grained_comet}
\end{table}

\subsection{LLM-as-a-judge Metrics Results}
Following ~\citet{sun-etal-2025-fine}, we complement automated metrics with LLM-as-a-judge metrics along multiple orthogonal dimensions, including fluency, content errors, and coherence errors. To avoid self-preference bias~\cite{chen-etal-2025-beyond}, we use Gemini-2.5-Flash for all systems. The results of News-Commentary Zh $\Rightarrow$ En are shown in Table~\ref{table:llm_results}. StarPO consistently achieves superior performance across all dimensions and model families, outperforming both standard CPO and strong baselines like Tower+ and GPT-4o.


\begin{table}[]
\centering
\small
\begin{NiceTabular}{lccc}
\toprule
& \bf Fluency ($\uparrow$) & \bf Content ($\downarrow$) & \bf Cohesion ($\downarrow$)  \\\midrule
\rowcolor{gray!15} 
\Block[c]{1-4}{\textsc{LLaMA-3.1-Instruct}}       \\
Base & 3.67 & 2.33 & 1.95  \\
+ SFT & 3.94 & 1.96 &  1.70    \\
\ + CPO & 3.97 & 1.90 & 1.58  \\
\ + StarPO& 3.99 & 1.40 & 1.27    \\ \midrule
\rowcolor{gray!15} 
\Block[c]{1-4}{\textsc{Qwen-2.5-7B-Instruct}}       \\
Base & 3.97 & 1.64 & 1.38  \\
+ SFT & 4.12 & 1.63 & 1.30 \\
+ CPO & \underline{4.40} & 1.39 & \underline{1.02} \\ 
+ StarPO & \bf 4.45 & \bf 1.16 & \bf 0.95  \\ \midrule
\rowcolor{gray!15} 
\Block[c]{1-4}{\textsc{Qwen-3-4B-Instruct}}       \\
Base & 4.39 & 1.46 & 1.13  \\
+ SFT & 4.51 & 1.44 & 1.11  \\
+ CPO & 4.58 & 1.23 & 0.91  \\ 
+ StarPO & 4.59 & 1.17 & 0.89  \\\midrule
\rowcolor{gray!15} 
\Block[c]{1-4}{\textsc{Other Systems}}       \\
Tower+ & 4.05 & 1.29 & 1.14  \\
GPT-4o & 4.18 & \underline{1.26} & 1.12  \\
Deepseek-R1 & 4.37 & 1.37 & 1.20 \\
\bottomrule
\end{NiceTabular}
\caption{LLM-as-a-judge evaluation results. Higher is better ($\uparrow$); lower is better ($\downarrow$).}
\label{table:llm_results}
\end{table}

\section{Related Work}

\subsection{Document-level Machine Translation}
LLM-based document-level machine translation generally falls into two paradigms: Doc2Sent (context-aware), which treats documents as sequences of sentence-level tasks, and Doc2Doc, which processes documents holistically.

In Doc2Sent style, training-free approaches rely on prompting strategies, such as context selection~\cite{wang-etal-2023-document-level,sia-duh-2023-context,moslem-etal-2023-adaptive,zhang-etal-2023-prompting,lee-etal-2025-testset,lippmann-etal-2025-context,cui-etal-2024-efficiently,peng-etal-2025-self,hu-etal-2025-source}, self-refinement~\cite{koneru-etal-2024-contextual,li-etal-2025-enhancing-large} and employing memory-based agents~\cite{wang-etal-2025-delta,guo-etal-2025-doc}. Fine-tuning relies on various data construction strategies~\cite{li-etal-2026-cross,lyu-etal-2024-dempt,wu-etal-2024-adapting,zhang-etal-2023-machine,stap-etal-2024-fine}, with the Tower series~\cite{alves-etal-2024-tower,rei-etal-2024-tower,rei-etal-2026-towerplus} being a representative case. Several studies further investigate the role and utilization of contextual information in context-aware translation~\cite{maka-etal-2025-train,maka-etal-2025-analyzing,mohammed-niculae-2025-context,choudhary-etal-2025-exploring,li-etal-2026-cross}.



Doc2Doc approaches aim for holistic translation through long-context training~\cite{pang-etal-2025-salute,li-etal-2026-enhancing}, iterative or agentic refinement~\cite{dong-etal-2025-two,li-etal-2025-enhancing-large,briakou-etal-2024-translating,wu-etal-2024-transagents}, and input optimization strategies like segmentation or knowledge fusion~\cite{hong-etal-2025-subdoctrans,liu-etal-2025-improving}. For evaluation, recent works~\cite{guo-etal-2025-auto,guo-etal-2025-align,domhan-zhu-2025-evaluation,wang-etal-2025-extending,steingrimsson-etal-2023-sentalign} predominantly rely on source-target sentence alignment to assess document quality.

\subsection{Reinforcement Learning for MT}
As references are not necessarily superior to LLM generations, RL~\cite{ouyang-etal-2022-rlhf} becomes essential for advancing MT. Recent research focus on training specialized reward models to guide this process~\cite{li-etal-2025-rival,feng-etal-2025-mt,ramos-etal-2026-finegrained,tan-monz-2025-remedy}.

RL approaches are generally categorized into online and offline methods. Online methods and reward-based methods, commonly use quality estimation models as reward models~\cite{he-etal-2024-improving,he-etal-2025-r1t1}. Such online frameworks is also compatible for training large reasoning models~\cite{feng-etal-2025-mt-r1,feng-etal-2026-mt3,wang-etal-2025-drt,wang-etal-2025-extrans,wang-etal-2026-deeptrans}. Typically, these methods apply various items into reward models. Conversely, offline methods like DPO~\cite{rafailov-etal-2023-dpo} and its variants~\cite{ethayarajh-etal-2024-kto,meng-etal-2024-simpo,xu-etal-2024-contrastive,xu-etal-2025-xalma,zeng-etal-2024-tim} rely on pre-curated preference datasets for stability. Various approaches are proposed to construct and leverage these datasets~\cite{agrawal-etal-2024-modeling,yang-etal-2024-direct,sun-etal-2025-enhancing-machine,cui-etal-2025-crpo,tang-etal-2025-mitigating,wang-etal-2026-beyond}.

\section{Conclusion}
To address structural misalignment in Doc2Doc translation, we introduce \textbf{STAR}, a metric for evaluating document-level structural fidelity, and \textbf{StarPO}, a preference optimization framework utilizing dynamic masking to target omissions and hallucinations. Experiments demonstrate StarPO enables compact models to surpass massive proprietary systems in translation quality while significantly improving token efficiency. This work establishes a robust paradigm for Doc2Doc translation without complex agentic workflows.

\section{Acknowledgments}

We thank the reviewers for their valuable and constructive feedback. This work was supported by Alibaba Group. Yichen Dong conducted this work during his internship at Alibaba Group.

\section*{Limitations}
First, in ``in-one-go'' Doc2Doc scenarios, establishing sentence-level alignment is an unavoidable prerequisite for calculating any fine-grained quality metric (e.g., d-COMET). Second, while enforcing 1-to-1 alignment effectively mitigates hallucinations, it imposes a structural rigidity that could theoretically discourage valid complex mappings in stylized texts, though our empirical results suggest this impact is minimal. Third, our experimental validation is currently concentrated on compact models (4B to 9B parameters) and high-to-medium resource languages. Validating the scalability of StarPO to larger architectures (e.g., 70B+) and low-resource languages remains a critical direction for future research. Finally, regarding data construction, we currently leverage proprietary models (e.g., GPT-4o) to augment candidate diversity. Although the selection of high-quality samples is strictly governed by our own STAR metric, this reliance on commercial APIs for initial generation currently prevents a fully end-to-end open-source pipeline. Future work aims to substitute this step with open-source alternatives, thereby enabling a completely offline-deployable training framework.

\bibliography{main}

@inproceedings{frohmann-etal-2024-segment,
    title = "Segment Any Text: A Universal Approach for Robust, Efficient and Adaptable Sentence Segmentation",
    author = "Frohmann, Markus  and
      Sterner, Igor  and
      Vuli{\'c}, Ivan  and
      Minixhofer, Benjamin  and
      Schedl, Markus",
    booktitle = "Proceedings of EMNLP",
    year = "2024",
    url = "https://aclanthology.org/2024.emnlp-main.665/",
    pages = "11908--11941"
}

@article{liu-zhu-2023-bertalign,
  title={Bertalign: Improved word embedding-based sentence alignment for Chinese--English parallel corpora of literary texts},
  author={Liu, Lei and Zhu, Min},
  journal={Digital Scholarship in the Humanities},
  pages={621--634},
  volume={38},
  year={2023},
  url = {https://doi.org/10.1093/llc/fqac089},
}

@inproceedings{
    xu-etal-2024-contrastive,
    title={Contrastive Preference Optimization: Pushing the Boundaries of {LLM} Performance in Machine Translation},
    author={Haoran Xu and Amr Sharaf and Yunmo Chen and Weiting Tan and Lingfeng Shen and Benjamin Van Durme and Kenton Murray and Young Jin Kim},
    booktitle={Proceedings of ICML},
    year={2024},
    url={https://openreview.net/forum?id=51iwkioZpn}
}

@inproceedings{
xu-etal-2025-xalma,
title={X-{ALMA}: Plug \& Play Modules and Adaptive Rejection for Quality Translation at Scale},
author={Haoran Xu and Kenton Murray and Philipp Koehn and Hieu Hoang and Akiko Eriguchi and Huda Khayrallah},
booktitle={Proceedings of ICLR},
year={2025},
url={https://openreview.net/forum?id=csbf1p8xUq}
}

@inproceedings{sun-etal-2025-fine,
    title = "Fine-Grained and Multi-Dimensional Metrics for Document-Level Machine Translation",
    author = "Sun, Yirong  and
      Zhu, Dawei  and
      Chen, Yanjun  and
      Xiao, Erjia  and
      Chen, Xinghao  and
      Shen, Xiaoyu",
    booktitle = "Proceedings of NAACL:HLT",
    year = "2025",
    url = "https://aclanthology.org/2025.naacl-srw.1/",
    pages = "1--17",

}

@inproceedings{dong-etal-2025-two,
    title = "Two Intermediate Translations Are Better Than One: Fine-tuning {LLM}s for Document-level Translation Refinement",
    author = "Dong, Yichen  and
      Lyu, Xinglin  and
      Li, Junhui  and
      Wei, Daimeng  and
      Zhang, Min  and
      Tao, Shimin  and
      Yang, Hao",
    booktitle = "Proceedings of ACL",
    year = "2025",
    url = "https://aclanthology.org/2025.acl-long.726/",
    pages = "14917--14933",
}

@inproceedings{liu-etal-2025-improving,
author = {Liu, Bin and Lyu, Xinglin and Li, Junhui and Wei, Daimeng and Zhang, Min and Tao, Shimin and Yang, Hao},
title = {Improving LLM-Based Document-Level MT with Multi-Knowledge Fusion},
year = {2025},
url = {https://doi.org/10.1007/978-981-95-3349-7_14},
booktitle = {Proceedings of NLPCC},
pages = {175–187},
}

@inproceedings{feng-etal-2025-mt-r1,
    title = "{MT}-R1-Zero: Advancing {LLM}-based Machine Translation via R1-Zero-like Reinforcement Learning",
    author = "Feng, Zhaopeng  and
      Cao, Shaosheng  and
      Ren, Jiahan  and
      Su, Jiayuan  and
      Chen, Ruizhe  and
      Zhang, Yan  and
      Wu, Jian  and
      Liu, Zuozhu",
    booktitle = "Findings of the EMNLP",
    year = "2025",
    url = "https://aclanthology.org/2025.findings-emnlp.1015/",
    pages = "18685--18702",

}

@article{he-etal-2025-r1t1,
  title={R1-t1: Fully incentivizing translation capability in llms via reasoning learning},
  author={He, Minggui and Liu, Yilun and Tao, Shimin and Luo, Yuanchang and Zeng, Hongyong and Su, Chang and Zhang, Li and Ma, Hongxia and Wei, Daimeng and Meng, Weibin and others},
  journal      = {CoRR},
  volume       = {abs/2502.19735},
  year         = {2025},
  url          = {https://doi.org/10.48550/arXiv.2502.19735},
}

@article{li-etal-2026-cross,
      title={Cross-Preference Learning for Sentence-Level and Context-Aware Machine Translation}, 
      author={Ying Li and Xinglin Lyu and Junhui Li and Jinlong Yang and Hengchao Shang and Min Zhang and Shimin Tao and Daimeng Wei},
  journal      = {CoRR},
  volume       = {abs/2603.25183},
  year={2026},
  url={https://arxiv.org/abs/2603.25183}, 
}

@inproceedings{wu-etal-2024-word,
    title = "Word Alignment as Preference for Machine Translation",
    author = "Wu, Qiyu  and
      Nagata, Masaaki  and
      Miao, Zhongtao  and
      Tsuruoka, Yoshimasa",
    booktitle = "Proceedings of EMNLP",
    year = "2024",
    url = "https://aclanthology.org/2024.emnlp-main.188/",
    doi = "10.18653/v1/2024.emnlp-main.188",
    pages = "3223--3239",
}

@inproceedings{wu-etal-2023-wspalign,
    title = "{WSPA}lign: Word Alignment Pre-training via Large-Scale Weakly Supervised Span Prediction",
    author = "Wu, Qiyu  and Nagata, Masaaki  and Tsuruoka, Yoshimasa",
    booktitle = "Proceedings of ACL",
    year = "2023",
    url = "https://aclanthology.org/2023.acl-long.621",
    pages = "11084--11099",
}

@article{zheng-etal-2025-gspo,
      title={Group Sequence Policy Optimization}, 
      author={Chujie Zheng and Shixuan Liu and Mingze Li and Xiong-Hui Chen and Bowen Yu and Chang Gao and Kai Dang and Yuqiong Liu and Rui Men and An Yang and Jingren Zhou and Junyang Lin},
  journal      = {CoRR},
  volume       = {abs/2507.18071},
  year         = {2025},
  url          = {https://doi.org/10.48550/arXiv.2507.18071},
}

@inproceedings{feng-etal-2026-mt3,
    title = "{MT}$^{3}$: A Synergistic Multi-Task {RL} Framework for Specializing {MLLM}s in Text Image Machine Translation",
    author = "Feng, Zhaopeng  and
      Liang, Yupu  and
      Cao, Shaosheng  and
      Su, Jiayuan  and
      Ren, Jiahan  and
      Zhou, Zhijie  and
      Huang, Wenxuan  and
      Wu, Jian  and
      Liu, Zuozhu",
    booktitle = "Proceedings of ACL",
    year = "2026",
    url = "https://aclanthology.org/2026.acl-long.460/",
    pages = "10140--10157",
}

@inproceedings{sia-duh-2023-context,
    title = "In-context Learning as Maintaining Coherency: A Study of On-the-fly Machine Translation Using Large Language Models",
    author = "Sia, Suzanna  and
      Duh, Kevin",
    booktitle = "Proceedings of Machine Translation Summit XIX, Vol. 1: Research Track",
    year = "2023",
    url = "https://aclanthology.org/2023.mtsummit-research.15/",
    pages = "173--185",
}

@inproceedings{cui-etal-2024-efficiently,
    title = "Efficiently Exploring Large Language Models for Document-Level Machine Translation with In-context Learning",
    author = "Cui, Menglong  and
      Du, Jiangcun  and
      Zhu, Shaolin  and
      Xiong, Deyi",
    booktitle = "Findings of ACL",
    year = "2024",
    url = "https://aclanthology.org/2024.findings-acl.646/",
    doi = "10.18653/v1/2024.findings-acl.646",
    pages = "10885--10897",
}

@inproceedings{wang-etal-2023-document-level,
    title = "Document-Level Machine Translation with Large Language Models",
    author = "Wang, Longyue  and
      Lyu, Chenyang  and
      Ji, Tianbo  and
      Zhang, Zhirui  and
      Yu, Dian  and
      Shi, Shuming  and
      Tu, Zhaopeng",
    booktitle = "Proceedings of EMNLP",
    year = "2023",
    url = "https://aclanthology.org/2023.emnlp-main.1036/",
    doi = "10.18653/v1/2023.emnlp-main.1036",
    pages = "16646--16661",
}

@inproceedings{wang-etal-2025-delta,
      title={DelTA: An Online Document-Level Translation Agent Based on Multi-Level Memory}, 
      author={Yutong Wang and Jiali Zeng and Xuebo Liu and Derek F. Wong and Fandong Meng and Jie Zhou and Min Zhang},
      year={2025},
      booktitle = "Proceedings of ICLR",
      url={https://openreview.net/forum?id=hoYFLRNbhc}, 
}

@inproceedings{guo-etal-2025-doc,
author = {Guo, Jiaxin and Luo, Yuanchang and Wei, Daimeng and Zhang, Ling and Li, Zongyao and Shang, Hengchao and Rao, Zhiqiang and Li, Shaojun and Yang, Jinlong and Wu, Zhanglin and Yang, Hao},
title = {Doc-Guided Sent2Sent++: A Sent2Sent++ Agent with Doc-Guided Memory for Document-Level Machine Translation},
year = {2025},
url = {https://doi.org/10.1007/978-981-95-3349-7_18},

booktitle = {Proceedings of NLPCC},
pages = {228–240},
}

@inproceedings{briakou-etal-2024-translating,
    title = "Translating Step-by-Step: Decomposing the Translation Process for Improved Translation Quality of Long-Form Texts",
    author = "Briakou, Eleftheria  and Luo, Jiaming  and Cherry, Colin  and Freitag, Markus",
    booktitle = "Proceedings of WMT",
    year = "2024",
    url = "https://aclanthology.org/2024.wmt-1.123/",
    doi = "10.18653/v1/2024.wmt-1.123",
    pages = "1301--1317",
}

@inproceedings{hu-etal-2025-source,
    title = "Source-primed Multi-turn Conversation Helps Large Language Models Translate Documents",
    author = "Hu, Hanxu  and
      Vamvas, Jannis  and
      Sennrich, Rico",
    booktitle = "Findings of EMNLP",
    year = "2025",
    url = "https://aclanthology.org/2025.findings-emnlp.1289/",
    doi = "10.18653/v1/2025.findings-emnlp.1289",
    pages = "23702--23712",
}

@article{pang-etal-2025-salute,
    title = "Salute the Classic: Revisiting Challenges of Machine Translation in the Age of Large Language Models",
    author = "Pang, Jianhui  and
      Ye, Fanghua  and
      Wong, Derek Fai  and
      Yu, Dian  and
      Shi, Shuming  and
      Tu, Zhaopeng  and
      Wang, Longyue",
    journal = "Transactions of the Association for Computational Linguistics",
    volume = "13",
    year = "2025",
    url = "https://aclanthology.org/2025.tacl-1.4/",
    doi = "10.1162/tacl_a_00730",
    pages = "73--95",

}

@article{wu-etal-2024-adapting,
  title={Adapting large language models for document-level machine translation},
  author={Wu, Minghao and Vu, Thuy-Trang and Qu, Lizhen and Foster, George and Haffari, Gholamreza},
  volume={abs/2401.06468},
  journal={CoRR},
  url = "https://arxiv.org/abs/2401.06468",
  year={2024}
}

@inproceedings{lyu-etal-2024-dempt,
    title = "{D}e{MPT}: Decoding-enhanced Multi-phase Prompt Tuning for Making {LLM}s Be Better Context-aware Translators",
    author = "Lyu, Xinglin  and Li, Junhui  and Zhao, Yanqing  and Zhang, Min  and Wei, Daimeng  and Tao, Shimin  and Yang, Hao  and Zhang, Min",
    booktitle = "Proceedings of EMNLP",
    year = "2024",
    url = "https://aclanthology.org/2024.emnlp-main.1131",
    pages = "20280--20295",
}

@article{alves-etal-2024-tower,
      title={Tower: An Open Multilingual Large Language Model for Translation-Related Tasks}, 
      author={Duarte M. Alves and José Pombal and Nuno M. Guerreiro and Pedro H. Martins and João Alves and Amin Farajian and Ben Peters and Ricardo Rei and Patrick Fernandes and Sweta Agrawal and Pierre Colombo and José G. C. de Souza and André F. T. Martins},
  journal      = {CoRR},
  volume       = {abs/2402.17733},
  year         = {2024},
  url          = {https://doi.org/10.48550/arXiv.2402.17733},
}

@inproceedings{rei-etal-2024-tower,
    title = "Tower v2: Unbabel-{IST} 2024 Submission for the General {MT} Shared Task",
    author = "Rei, Ricardo  and
      Pombal, Jose  and
      Guerreiro, Nuno M.  and
      Alves, Jo{\~a}o  and
      Martins, Pedro Henrique  and
      Fernandes, Patrick  and
      Wu, Helena  and
      Vaz, Tania  and
      Alves, Duarte  and
      Farajian, Amin  and
      Agrawal, Sweta  and
      Farinhas, Antonio  and
      C. De Souza, Jos{\'e} G.  and
      Martins, Andr{\'e}",
    booktitle = "Proceedings of WMT",
    year = "2024",
    url = "https://aclanthology.org/2024.wmt-1.12/",
    doi = "10.18653/v1/2024.wmt-1.12",
    pages = "185--204",

}

@inproceedings{rafailov-etal-2023-dpo,
	title = {Direct {Preference} {Optimization}: {Your} {Language} {Model} is {Secretly} a {Reward} {Model}},
	url = {https://proceedings.neurips.cc/paper_files/paper/2023/file/a85b405ed65c6477a4fe8302b5e06ce7-Paper-Conference.pdf},
	booktitle = {Proceedings of NIPS},
	author = {Rafailov, Rafael and Sharma, Archit and Mitchell, Eric and Manning, Christopher D and Ermon, Stefano and Finn, Chelsea},
	year = {2023},
	pages = {53728--53741},
}

@article{shao-etal-2024-deepseekmath,
      title={DeepSeekMath: Pushing the Limits of Mathematical Reasoning in Open Language Models}, 
      author={Zhihong Shao and Peiyi Wang and Qihao Zhu and Runxin Xu and Junxiao Song and Xiao Bi and Haowei Zhang and Mingchuan Zhang and Y. K. Li and Y. Wu and Daya Guo}, 
  journal      = {CoRR},
  volume       = {abs/2402.03300},
  year         = {2024},
  url          = {https://doi.org/10.48550/arXiv.2402.03300},
}

@inproceedings{zouhar-etal-2024-pitfalls,
    title = "Pitfalls and Outlooks in Using {COMET}",
    author = "Zouhar, Vil{\'e}m  and Chen, Pinzhen  and Lam, Tsz Kin  and Moghe, Nikita  and Haddow, Barry",
    booktitle = "Proceedings of WMT",
    year = "2024",
    url = "https://aclanthology.org/2024.wmt-1.121/",
    pages = "1272--1288",
}

@inproceedings{zebaze-etal-2025-context,
    title = "In-Context Example Selection via Similarity Search Improves Low-Resource Machine Translation",
    author = "Zebaze, Armel Randy  and
      Sagot, Beno{\^i}t  and
      Bawden, Rachel",
    booktitle = "Findings of NAACL",
    year = "2025",
    url = "https://aclanthology.org/2025.findings-naacl.68/",
    doi = "10.18653/v1/2025.findings-naacl.68",
    pages = "1222--1252",
}

@inproceedings{wang-etal-2024-benchmarking,
    title = "Benchmarking and Improving Long-Text Translation with Large Language Models",
    author = "Wang, Longyue  and
      Du, Zefeng  and
      Jiao, Wenxiang  and
      Lyu, Chenyang  and
      Pang, Jianhui  and
      Cui, Leyang  and
      Song, Kaiqiang  and
      Wong, Derek  and
      Shi, Shuming  and
      Tu, Zhaopeng",
    booktitle = "Findings ACL",
    year = "2024",
    url = "https://aclanthology.org/2024.findings-acl.428/",
    doi = "10.18653/v1/2024.findings-acl.428",
    pages = "7175--7187",
}

@inproceedings{moslem-etal-2023-adaptive,
    title = "Adaptive Machine Translation with Large Language Models",
    author = "Moslem, Yasmin  and
      Haque, Rejwanul  and
      Kelleher, John D.  and
      Way, Andy",
    booktitle = "Proceedings of EAMT",
    year = "2023",
    url = "https://aclanthology.org/2023.eamt-1.22/",
    pages = "227--237",
}

@inproceedings{zhang-etal-2023-prompting,
  title={Prompting large language model for machine translation: A case study},
  author={Zhang, Biao and Haddow, Barry and Birch, Alexandra},
  booktitle={Proceedings of ICML},
  pages={41092--41110},
  year={2023},
  url={https://openreview.net/pdf?id=yWl0agiI0y}
}

@inproceedings{koneru-etal-2024-contextual,
    title = "Contextual Refinement of Translations: Large Language Models for Sentence and Document-Level Post-Editing",
    author = "Koneru, Sai  and
      Exel, Miriam  and
      Huck, Matthias  and
      Niehues, Jan",
    booktitle = "Proceedings of NAACL:HLT",
    year = "2024",
    url = "https://aclanthology.org/2024.naacl-long.148/",
    doi = "10.18653/v1/2024.naacl-long.148",
    pages = "2711--2725",
}

@inproceedings{wu-etal-2024-transagents,
    title = "{T}rans{A}gents: Build Your Translation Company with Language Agents",
    author = "Wu, Minghao  and
      Xu, Jiahao  and
      Wang, Longyue",
    booktitle = "Proceedings of EMNLP: System Demonstrations",
    year = "2024",
    url = "https://aclanthology.org/2024.emnlp-demo.14/",
    doi = "10.18653/v1/2024.emnlp-demo.14",
    pages = "131--141",
}

@inproceedings{zhang-etal-2023-machine,
    title = "Machine Translation with Large Language Models: Prompting, Few-shot Learning, and Fine-tuning with {QL}o{RA}",
    author = "Zhang, Xuan  and
      Rajabi, Navid  and
      Duh, Kevin  and
      Koehn, Philipp",
    booktitle = "Proceedings of WMT",
    year = "2023",
    url = "https://aclanthology.org/2023.wmt-1.43/",
    doi = "10.18653/v1/2023.wmt-1.43",
    pages = "468--481",
}

@inproceedings{maka-etal-2025-train,
    title = "You Are What You Train: Effects of Data Composition on Training Context-aware Machine Translation Models",
    author = "M{\k{a}}ka, Pawe{\l}  and
      Semerci, Yusuf Can  and
      Scholtes, Jan  and
      Spanakis, Gerasimos",
    booktitle = "Proceedings of EMNLP",
    year = "2025",
    url = "https://aclanthology.org/2025.emnlp-main.1394/",
    doi = "10.18653/v1/2025.emnlp-main.1394",
    pages = "27402--27425",
}

@inproceedings{mohammed-niculae-2025-context,
    title = "Context-Aware or Context-Insensitive? Assessing {LLM}s' Performance in Document-Level Translation",
    author = "Mohammed, Wafaa  and
      Niculae, Vlad",
    booktitle = "Proceedings of Machine Translation Summit XX: Volume 1",
    year = "2025",
    url = "https://aclanthology.org/2025.mtsummit-1.10/",
    pages = "126--137",
}

@article{li-etal-2026-enhancing,
title = {Enhancing document-level translation of large language model via translation mixed instructions},
journal = {Neurocomputing},
volume = {664},
pages = {132041},
year = {2026},
issn = {0925-2312},
doi = {https://doi.org/10.1016/j.neucom.2025.132041},
url = {https://www.sciencedirect.com/science/article/pii/S0925231225027134},
author = {Yachao Li and Junhui Li and Jing Jiang and Min Zhang},
}

@inproceedings{li-etal-2025-enhancing-large,
    title = "Enhancing Large Language Models for Document-Level Translation Post-Editing Using Monolingual Data",
    author = "Li, Zongyao  and
      Rao, Zhiqiang  and
      Shang, Hengchao  and
      Guo, Jiaxin  and
      Li, Shaojun  and
      Wei, Daimeng  and
      Yang, Hao",
    booktitle = "Proceedings of COLING",
    year = "2025",
    url = "https://aclanthology.org/2025.coling-main.591/",
    pages = "8830--8840",
}

@inproceedings{domhan-zhu-2025-evaluation,
    title = "Same evaluation, more tokens: On the effect of input length for machine translation evaluation using Large Language Models",
    author = "Domhan, Tobias  and
      Zhu, Dawei",
    booktitle = "Proceedings of EMNLP",
    year = "2025",
    url = "https://aclanthology.org/2025.emnlp-main.402/",
    doi = "10.18653/v1/2025.emnlp-main.402",
    pages = "7940--7958",
}

@inproceedings{wang-etal-2025-extending,
    title = "Extending Automatic Machine Translation Evaluation to Book-Length Documents",
    author = "Wang, Kuang-Da  and
      Ding, Shuoyang  and
      Yang, Chao-Han Huck  and
      Hsieh, Ping-Chun  and
      Peng, Wen-Chih  and
      Lavrukhin, Vitaly  and
      Ginsburg, Boris",
    booktitle = "Proceedings of EMNLP",
    year = "2025",
    url = "https://aclanthology.org/2025.emnlp-main.1645/",
    doi = "10.18653/v1/2025.emnlp-main.1645",
    pages = "32311--32327",
}

@inproceedings{li-etal-2025-rival,
    title = "{RIVAL}: Reinforcement Learning with Iterative and Adversarial Optimization for Machine Translation",
    author = "Li, Tianjiao  and
      Yu, Mengran  and
      Shi, Chenyu  and
      Zhao, Yanjun  and
      Liu, Xiaojing  and
      Zhang, Qi  and
      Huang, Xuanjing  and
      Zhang, Qiang  and
      Wang, Jiayin",
    booktitle = "Findings of EMNLP 2025",
    year = "2025",
    url = "https://aclanthology.org/2025.findings-emnlp.166/",
    doi = "10.18653/v1/2025.findings-emnlp.166",
    pages = "3064--3079",
}

@inproceedings{feng-etal-2025-mt,
    title = "{MT}-{R}eward{T}ree: A Comprehensive Framework for Advancing {LLM}-Based Machine Translation via Reward Modeling",
    author = "Feng, Zhaopeng  and
      Ren, Jiahan  and
      Su, Jiayuan  and
      Zheng, Jiamei  and
      Wang, Hongwei  and
      Liu, Zuozhu",
    booktitle = "Findings of EMNLP",
    year = "2025",
    url = "https://aclanthology.org/2025.findings-emnlp.1007/",
    doi = "10.18653/v1/2025.findings-emnlp.1007",
    pages = "18556--18567",
}

@inproceedings{hong-etal-2025-subdoctrans,
    title = "{S}ub{D}oc{T}rans: Enhancing Document-level Machine Translation with Plug-and-play Multi-granularity Knowledge Augmentation",
    author = "Hong, Hanghai  and
      Xie, Yibo  and
      Zheng, Jiawei  and
      Wang, Xiaoli",
    booktitle = "Findings of EMNLP",
    year = "2025",
    url = "https://aclanthology.org/2025.findings-emnlp.782/",
    doi = "10.18653/v1/2025.findings-emnlp.782",
    pages = "14490--14506",
}

@inproceedings{choudhary-etal-2025-exploring,
    title = "Exploring Context Strategies in {LLM}s for Discourse-Aware Machine Translation",
    author = "Choudhary, Ritvik  and
      Hida, Rem  and
      Hamada, Masaki  and
      Futami, Hayato  and
      Sekiya, Toshiyuki",
    booktitle = "Findings of EMNLP",
    year = "2025",
    url = "https://aclanthology.org/2025.findings-emnlp.1324/",
    doi = "10.18653/v1/2025.findings-emnlp.1324",
    pages = "24382--24391",
}

@inproceedings{zhao-etal-2025-swift,
      title={SWIFT:A Scalable lightWeight Infrastructure for Fine-Tuning},
      author={Yuze Zhao and Jintao Huang and Jinghan Hu and Xingjun Wang and Yunlin Mao and Daoze Zhang and Zeyinzi Jiang and Zhikai Wu and Baole Ai and Ang Wang and Wenmeng Zhou and Yingda Chen},
  booktitle    = {Proceedings of AAAI},
  pages        = {29733--29735},
  year         = {2025},
  url = {https://doi.org/10.1609/aaai.v39i28.35383},
}

@inproceedings{ouyang-etal-2022-rlhf,
  author       = {Long Ouyang and
                  Jeffrey Wu and
                  Xu Jiang and
                  Diogo Almeida and
                  Carroll L. Wainwright and
                  Pamela Mishkin and
                  Chong Zhang and
                  Sandhini Agarwal and
                  Katarina Slama and
                  Alex Ray and
                  John Schulman and
                  Jacob Hilton and
                  Fraser Kelton and
                  Luke Miller and
                  Maddie Simens and
                  Amanda Askell and
                  Peter Welinder and
                  Paul F. Christiano and
                  Jan Leike and
                  Ryan Lowe},
  title        = {Training language models to follow instructions with human feedback},
  booktitle    = {Proceedings of NIPS},
  year         = {2022},
  url          = {http://papers.nips.cc/paper\_files/paper/2022/hash/b1efde53be364a73914f58805a001731-Abstract-Conference.html},

}

@inproceedings{ethayarajh-etal-2024-kto,
  title = 	 {Model Alignment as Prospect Theoretic Optimization},
  author =       {Ethayarajh, Kawin and Xu, Winnie and Muennighoff, Niklas and Jurafsky, Dan and Kiela, Douwe},
  booktitle = 	 {Proceedings of ICML},
  pages = 	 {12634--12651},
  year = 	 {2024},
  url = 	 {https://proceedings.mlr.press/v235/ethayarajh24a.html},
}

@inproceedings{meng-etal-2024-simpo,
  title={Simpo: Simple preference optimization with a reference-free reward},
  author={Meng, Yu and Xia, Mengzhou and Chen, Danqi},
  booktitle={Proceedings of NIPS},
  pages={124198--124235},
  year={2024},
  url={http://papers.nips.cc/paper\_files/paper/2024/hash/e099c1c9699814af0be873a175361713-Abstract-Conference.html}
}

@inproceedings{sun-etal-2025-enhancing-machine,
    title = "Enhancing Machine Translation with Self-Supervised Preference Data",
    author = "Sun, Haoxiang  and
      Gao, Ruize  and
      Zhang, Pei  and
      Yang, Baosong  and
      Wang, Rui",
    booktitle = "Proceedings of ACL",
    year = "2025",
    url = "https://aclanthology.org/2025.acl-long.1165/",
    doi = "10.18653/v1/2025.acl-long.1165",
    pages = "23916--23934",
}

@inproceedings{zeng-etal-2024-tim,
 author={Jiali Zeng and Fandong Meng and Yongjing Yin and Jie Zhou},
  title        = {Teaching Large Language Models to Translate with Comparison},
  booktitle    = {Proceedings of AAAI },
  pages        = {19488--19496},
  year         = {2024},
  url          = {https://doi.org/10.1609/aaai.v38i17.29920},
}

@inproceedings{cui-etal-2025-crpo,
    title = "{CRPO}: Confidence-Reward Driven Preference Optimization for Machine Translation",
    author = "Cui, Guofeng  and
      Wang, Pichao  and
      Liu, Yang  and
      Ke, Zemian  and
      Liu, Zhu  and
      Bhat, Vimal",
    booktitle = "Findings of ACL",
    year = "2025",
    url = "https://aclanthology.org/2025.findings-acl.31/",
    doi = "10.18653/v1/2025.findings-acl.31",
    pages = "560--574",
}

@inproceedings{he-etal-2024-improving,
    title = "Improving Machine Translation with Human Feedback: An Exploration of Quality Estimation as a Reward Model",
    author = "He, Zhiwei  and
      Wang, Xing  and
      Jiao, Wenxiang  and
      Zhang, Zhuosheng  and
      Wang, Rui  and
      Shi, Shuming  and
      Tu, Zhaopeng",
    booktitle = "Proceedings of NAACL:HLT",
    year = "2024",
    url = "https://aclanthology.org/2024.naacl-long.451/",
    doi = "10.18653/v1/2024.naacl-long.451",
    pages = "8164--8180",
}

@inproceedings{tang-etal-2025-mitigating,
    title = "Mitigating Hallucinated Translations in Large Language Models with Hallucination-focused Preference Optimization",
    author = "Tang, Zilu  and
      Chatterjee, Rajen  and
      Garg, Sarthak",
    booktitle = "Proceedings of NAACL:HLT",
    year = "2025",
    url = "https://aclanthology.org/2025.naacl-long.175/",
    doi = "10.18653/v1/2025.naacl-long.175",
    pages = "3410--3433",
}

@inproceedings{lee-etal-2025-testset,
    title = "A Testset for Context-Aware {LLM} Translation in {K}orean-to-{E}nglish Discourse Level Translation",
    author = "Lee, Minjae  and
      Noh, Youngbin  and
      Lee, Seung Jin",
    booktitle = "Proceedings of COLING",
    year = "2025",
    url = "https://aclanthology.org/2025.coling-main.110/",
    pages = "1632--1646",
}

@inproceedings{maka-etal-2025-analyzing,
    title = "Analyzing the Attention Heads for Pronoun Disambiguation in Context-aware Machine Translation Models",
    author = "M{\k{a}}ka, Pawe{\l}  and
      Semerci, Yusuf Can  and
      Scholtes, Jan  and
      Spanakis, Gerasimos",
    booktitle = "Proceedings of COLING",
    year = "2025",
    url = "https://aclanthology.org/2025.coling-main.424/",
    pages = "6348--6377",
}

@inproceedings{lippmann-etal-2025-context,
    title = "Context-Informed Machine Translation of Manga using Multimodal Large Language Models",
    author = "Lippmann, Philip  and
      Skublicki, Konrad  and
      Tanner, Joshua  and
      Ishiwatari, Shonosuke  and
      Yang, Jie",
    booktitle = "Proceedings of COLING",
    year = "2025",
    url = "https://aclanthology.org/2025.coling-main.232/",
    pages = "3444--3464",
}

@inproceedings{agrawal-etal-2024-modeling,
    title = "Modeling User Preferences with Automatic Metrics: Creating a High-Quality Preference Dataset for Machine Translation",
    author = "Agrawal, Sweta  and
      De Souza, Jos{\'e} G. C.  and
      Rei, Ricardo  and
      Farinhas, Ant{\'o}nio  and
      Faria, Gon{\c{c}}alo  and
      Fernandes, Patrick  and
      Guerreiro, Nuno M  and
      Martins, Andre",
    booktitle = "Proceedings of EMNLP",
    month = nov,
    year = "2024",
    url = "https://aclanthology.org/2024.emnlp-main.803/",
    pages = "14503--14519",
}

@article{ramos-etal-2026-finegrained,
    author = {Ramos, Miguel Moura and Almeida, Tomás and Vareta, Daniel and Azevedo, Filipe and Agrawal, Sweta and Fernandes, Patrick and Martins, André F. T.},
    title = {Fine-Grained Reward Optimization for Machine Translation using Error
                    Severity Mappings},
    journal = {Transactions of the Association for Computational Linguistics},
    volume = {14},
    pages = {733-754},
    year = {2026},
    month = {05},
    url = {https://doi.org/10.1162/TACL.a.646},
}

@article{openai-etal-gpt4o-2024,
  author       = {OpenAI},
  title        = {{GPT-4o System Card}},
  journal      = {CoRR},
  volume       = {abs/2410.21276},
  year         = {2024},
  url          = {https://arxiv.org/abs/2410.21276},
}

@inproceedings{wang-etal-2026-beyond,
  author       = {Hao Wang and
                  Linlong Xu and
                  Heng Liu and
                  Yangyang Liu and
                  Xiaohu Zhao and
                  Bo Zeng and
                  Liangying Shao and
                  Yichen Dong and
                  Xinwei Wu and
                  Jiang Zhou and
                  Tianyu Dong and
                  Xiangxiang Zeng and
                  Longyue Wang and
                  Weihua Luo},
  title        = {M{\({^2}\)}PO: Multi-Perspective Multi-Pair Preference Optimization
                  for Machine Translation},
  booktitle    = {Proceedings of ACL},
  pages        = {10315--10336},
  year         = {2026},
  url          = {https://doi.org/10.18653/v1/2026.acl-long.469},
}

@article{guo-etal-2025-align,
      title={Align-then-Slide: A complete evaluation framework for Ultra-Long Document-Level Machine Translation}, 
      author={Jiaxin Guo and Daimeng Wei and Yuanchang Luo and Xiaoyu Chen and Zhanglin Wu and Huan Yang and Hengchao Shang and Zongyao Li and Zhiqiang Rao and Jinlong Yang and Hao Yang},
      journal      = {CoRR},
  volume       = {abs/2509.03809},
  year         = {2025},
  url          = {https://doi.org/10.48550/arXiv.2509.03809},
}

@inproceedings{tan-monz-2025-remedy,
    title = "{R}e{M}edy: Learning Machine Translation Evaluation from Human Preferences with Reward Modeling",
    author = "Tan, Shaomu  and
      Monz, Christof",
    booktitle = "Proceedings of EMNLP",
    year = "2025",
    url = "https://aclanthology.org/2025.emnlp-main.217/",
    doi = "10.18653/v1/2025.emnlp-main.217",
    pages = "4370--4387",
    ISBN = "979-8-89176-332-6",
}

@inproceedings{hu-etal-2021-lora,
  title={Lora: Low-rank adaptation of large language models},
  author={Hu, Edward J and Shen, Yelong and Wallis, Phillip and Allen-Zhu, Zeyuan and Li, Yuanzhi and Wang, Shean and Wang, Lu and Chen, Weizhu},
  booktitle = "Proceedings of ICLR",
  url={https://openreview.net/forum?id=nZeVKeeFYf9},
  year={2021}
}

@article{team-etal-2024-llama3,
        author       = {Llama Team},
  title        = {The Llama 3 Herd of Models},
  journal      = {CoRR},
  volume       = {abs/2407.21783},
  year         = {2024},
  url          = {https://doi.org/10.48550/arXiv.2407.21783}
}

@article{qwen2.5,
    title = {Qwen2.5: A Party of Foundation Models},
    url = {https://qwenlm.github.io/blog/qwen2.5/},
    author = {Qwen Team},
    month = {September},
    year = {2024}
}

@inproceedings{akter-etal-2026-detect,
    title = "Detecting Proxy Gaming in {RL} and {LLM} Alignment via Evaluator Stress Tests",
    author = "Akter, Sanjeda  and
      Shihab, Ibne Farabi  and
      Sharma, Anuj",
    editor = "Liakata, Maria  and
      Moreira, Viviane P.  and
      Zhang, Jiajun  and
      Jurgens, David",
    booktitle = "Findings of ACL",
    year = "2026",
    url = "https://aclanthology.org/2026.findings-acl.513/",
    pages = "10554--10583",
}

@inproceedings{rei-etal-2026-towerplus,
          title = "{TOWER}+: Bridging Generality and Translation Specialization in Multilingual {LLM}s",
    author = "Rei, Ricardo  and
      Guerreiro, Nuno M  and
      Pombal, Jos{\'e}  and
      Alves, Jo{\~a}o  and
      Farajian, Amin  and
      Teixeirinha, Pedro  and
      Martins, Andre",
    booktitle = "Proceedings of ACL",
    year = "2026",
    url = "https://aclanthology.org/2026.acl-long.1366/",
    doi = "10.18653/v1/2026.acl-long.1366",
    pages = "29614--29635",
}

@inproceedings{hong-etal-2024-orpo,
      title={ORPO: Monolithic Preference Optimization without Reference Model}, 
      author={Jiwoo Hong and Noah Lee and James Thorne},
  booktitle    = {Proceedings of EMNLP},
  pages        = {11170--11189},
  year         = {2024},
  url          = {https://doi.org/10.18653/v1/2024.emnlp-main.626},
}

@inproceedings{wang-etal-2025-drt,
    title = "{DRT}: Deep Reasoning Translation via Long Chain-of-Thought",
    author = "Wang, Jiaan  and
      Meng, Fandong  and
      Liang, Yunlong  and
      Zhou, Jie",
    booktitle = "Findings of ACL 2025",
    month = jul,
    year = "2025",
    url = "https://aclanthology.org/2025.findings-acl.351/",
    doi = "10.18653/v1/2025.findings-acl.351",
    pages = "6770--6782",
}

@inproceedings{vernikos-etal-2022-embarrassingly,
    title = "Embarrassingly Easy Document-Level {MT} Metrics: How to Convert Any Pretrained Metric into a Document-Level Metric",
    author = "Vernikos, Giorgos  and Thompson, Brian  and Mathur, Prashant  and Federico, Marcello",
    booktitle = "Proceedings of WMT",
    year = "2022",
    url = "https://aclanthology.org/2022.wmt-1.6",
    pages = "118--128",
}

@inproceedings{chen-etal-2025-beyond,
    title = "Beyond the Surface: Measuring Self-Preference in {LLM} Judgments",
    author = "Chen, Zhi-Yuan  and
      Wang, Hao  and
      Zhang, Xinyu  and
      Hu, Enrui  and
      Lin, Yankai",
    booktitle = "Proceedings of EMNLP",
    month = nov,
    year = "2025",
    address = "Suzhou, China",
    url = "https://aclanthology.org/2025.emnlp-main.86/",
    doi = "10.18653/v1/2025.emnlp-main.86",
    pages = "1653--1672",

}

@article{deepseek-2025-deepseek,
  author       = {DeepSeek{-}AI},
  title        = {DeepSeek-R1: Incentivizing Reasoning Capability in LLMs via Reinforcement
                  Learning},
  journal      = {CoRR},
  volume       = {abs/2501.12948},
  year         = {2025},
  url          = {https://doi.org/10.48550/arXiv.2501.12948}
}

@inproceedings{chen-etal-2024-m3,
    title = "{M}3-Embedding: Multi-Linguality, Multi-Functionality, Multi-Granularity Text Embeddings Through Self-Knowledge Distillation",
    author = "Chen, Jianlyu  and
      Xiao, Shitao  and
      Zhang, Peitian  and
      Luo, Kun  and
      Lian, Defu  and
      Liu, Zheng",
    booktitle = "Findings of ACL",
    year = "2024",
    url = "https://aclanthology.org/2024.findings-acl.137/",
    pages = "2318--2335",

}

@inproceedings{guerreiro-etal-2023-looking,
    title = "Looking for a Needle in a Haystack: A Comprehensive Study of Hallucinations in Neural Machine Translation",
    author = "Guerreiro, Nuno M.  and
      Voita, Elena  and
      Martins, Andr{\'e}",
    booktitle = "Proceedings of EACL",
    year = "2023",
    url = "https://aclanthology.org/2023.eacl-main.75/",
    doi = "10.18653/v1/2023.eacl-main.75",
    pages = "1059--1075",
}

@inproceedings{shao-etal-2024-understanding,
    title = "Understanding and Addressing the Under-Translation Problem from the Perspective of Decoding Objective",
    author = "Shao, Chenze  and
      Meng, Fandong  and
      Zeng, Jiali  and
      Zhou, Jie",
    editor = "Ku, Lun-Wei  and
      Martins, Andre  and
      Srikumar, Vivek",
    booktitle = "Proceedings of ACL",
    year = "2024",
    url = "https://aclanthology.org/2024.acl-long.209/",
    doi = "10.18653/v1/2024.acl-long.209",
    pages = "3800--3814",
}

@inproceedings{karpinska-iyyer-2023-large,
  title={Large Language Models Effectively Leverage Document-level Context for Literary Translation, but Critical Errors Persist},
  author={Marzena Karpinska and Mohit Iyyer},
  booktitle={Proceedings of WMT},
  pages={419-451},
  year={2023},
  url = {https://aclanthology.org/2023.wmt-1.41/}
}

@article{kocmi-etal-2024-preliminary,
      title={Preliminary WMT24 Ranking of General MT Systems and LLMs}, 
      author={Tom Kocmi and Eleftherios Avramidis and Rachel Bawden and Ondrej Bojar and Anton Dvorkovich and Christian Federmann and Mark Fishel and Markus Freitag and Thamme Gowda and Roman Grundkiewicz and Barry Haddow and Marzena Karpinska and Philipp Koehn and Benjamin Marie and Kenton Murray and Masaaki Nagata and Martin Popel and Maja Popovic and Mariya Shmatova and Steinþór Steingrímsson and Vilém Zouhar},
      year={2024},
      journal = {CoRR},    
      volume={abs/2407.19884},
      archivePrefix={arXiv},
      primaryClass={cs.CL},
      url={https://arxiv.org/abs/2407.19884}, 
}

@inproceedings{rei-etal-2022-comet,
    title = "{COMET}-22: Unbabel-{IST} 2022 Submission for the Metrics Shared Task",
    author = "Rei, Ricardo  and
      C. de Souza, Jos{\'e} G.  and
      Alves, Duarte  and
      Zerva, Chrysoula  and
      Farinha, Ana C  and
      Glushkova, Taisiya  and
      Lavie, Alon  and
      Coheur, Luisa  and
      Martins, Andr{\'e} F. T.",

    booktitle = "Proceedings of WMT",
    year = "2022",
    url = "https://aclanthology.org/2022.wmt-1.52/",
    pages = "578--585",
}

@article{liu-etal-2024-lost,
    title = "Lost in the Middle: How Language Models Use Long Contexts",
    author = "Liu, Nelson F.  and
      Lin, Kevin  and
      Hewitt, John  and
      Paranjape, Ashwin  and
      Bevilacqua, Michele  and
      Petroni, Fabio  and
      Liang, Percy",
    journal = "Transactions of the Association for Computational Linguistics",
    volume = "12",
    year = "2024",
    address = "Cambridge, MA",
    publisher = "MIT Press",
    url = "https://aclanthology.org/2024.tacl-1.9/",
    doi = "10.1162/tacl_a_00638",
    pages = "157--173",
}

@inproceedings{steingrimsson-etal-2023-sentalign,
    title = "{S}ent{A}lign: Accurate and Scalable Sentence Alignment",
    author = "Steingrimsson, Steinthor  and
      Loftsson, Hrafn  and
      Way, Andy",
    booktitle = "Proceedings of EMNLP: System Demonstrations",
    year = "2023",
    url = "https://aclanthology.org/2023.emnlp-demo.22/",
    doi = "10.18653/v1/2023.emnlp-demo.22",
    pages = "256--263",
}

@inproceedings{kocmi-etal-2024-navigating,
    title = "Navigating the Metrics Maze: Reconciling Score Magnitudes and Accuracies",
    author = "Kocmi, Tom  and
      Zouhar, Vil{\'e}m  and
      Federmann, Christian  and
      Post, Matt",
    booktitle = "Proceedings of ACL",
    year = "2024",
    url = "https://aclanthology.org/2024.acl-long.110/",
    pages = "1999--2014",
}

@inproceedings{gu-etal-2025-maskdpo,
title={Mask-{DPO}: Generalizable Fine-grained Factuality Alignment of {LLM}s},
author={Yuzhe Gu and Wenwei Zhang and Chengqi Lyu and Dahua Lin and Kai Chen},
booktitle={Proceedings of ICLR},
year={2025},
url={https://openreview.net/forum?id=d2H1oTNITn}
}

@inproceedings{peng-etal-2025-investigating,
    title = "Investigating Length Issues in Document-level Machine Translation",
    author = "Peng, Ziqian  and
      Bawden, Rachel  and
      Yvon, Fran{\c{c}}ois",
    booktitle = "Proceedings of Machine Translation Summit XX: Volume 1",
    year = "2025",
    url = "https://aclanthology.org/2025.mtsummit-1.3/",
    pages = "4--23",

}

@article{guo-etal-2025-auto,
      title={Automatic Evaluation Metrics for Document-level Translation: Overview, Challenges and Trends}, 
      author={Jiaxin Guo and Xiaoyu Chen and Zhiqiang Rao and Jinlong Yang and Zongyao Li and Hengchao Shang and Daimeng Wei and Hao Yang},
      year={2025},
      journal = {CoRR},    
      volume={abs/2504.14804},
      url={https://arxiv.org/abs/2504.14804}, 
}

@article{wang-etal-2026-deeptrans,
    title = "{D}eep{T}rans: Deep Reasoning Translation via Reinforcement Learning",
    author = "Wang, Jiaan  and
      Meng, Fandong  and
      Zhou, Jie",
    journal = "Transactions of the Association for Computational Linguistics",
    volume = "14",
    year = "2026",
    address = "Cambridge, MA",
    publisher = "MIT Press",
    url = "https://aclanthology.org/2026.tacl-1.3/",
    doi = "10.1162/tacl.a.65",
    pages = "47--63",
}

@article{wang-etal-2025-extrans,
  author       = {Jiaan Wang and
                  Fandong Meng and
                  Jie Zhou},
  title        = {ExTrans: Multilingual Deep Reasoning Translation via Exemplar-Enhanced
                  Reinforcement Learning},
  journal      = {CoRR},
  volume       = {abs/2505.12996},
  year         = {2025},
  url          = {https://doi.org/10.48550/arXiv.2505.12996},
}

@inproceedings{stap-etal-2024-fine,
    title = "The Fine-Tuning Paradox: Boosting Translation Quality Without Sacrificing {LLM} Abilities",
    author = "Stap, David  and
      Hasler, Eva  and
      Byrne, Bill  and
      Monz, Christof  and
      Tran, Ke",
    booktitle = "Proceedings of ACL",
    year = "2024",
    url = "https://aclanthology.org/2024.acl-long.336/",
    pages = "6189--6206",
}

@inproceedings{peng-etal-2025-self,
    title = "Self-Retrieval from Distant Contexts for Document-Level Machine Translation",
    author = "Peng, Ziqian  and
      Bawden, Rachel  and
      Yvon, Fran{\c{c}}ois",
    booktitle = "Proceedings of WMT",
    year = "2025",
    url = "https://aclanthology.org/2025.wmt-1.13/",
    pages = "220--240",
}

@inproceedings{yang-etal-2024-direct,
    title = "Direct Preference Optimization for Neural Machine Translation with Minimum {B}ayes Risk Decoding",
    author = "Yang, Guangyu  and
      Chen, Jinghong  and
      Lin, Weizhe  and
      Byrne, Bill",
    booktitle = "Proceedings of NAACL:HLT (Volume 2: Short Papers)",
    year = "2024",
    url = "https://aclanthology.org/2024.naacl-short.34/",
    pages = "391--398",
}

@article{sun-etal-2024-hunyuan,
      title={Hunyuan-Large: An Open-Source MoE Model with 52 Billion Activated Parameters by Tencent}, 
      author={Xingwu Sun and Yanfeng Chen and Yiqing Huang and Ruobing Xie and Jiaqi Zhu and Kai Zhang and Shuaipeng Li and Zhen Yang and Jonny Han and Xiaobo Shu and Jiahao Bu and Zhongzhi Chen and Xuemeng Huang and Fengzong Lian and Saiyong Yang and Jianfeng Yan and Yuyuan Zeng and Xiaoqin Ren and Chao Yu and Lulu Wu and Yue Mao and Tao Yang and Suncong Zheng and Kan Wu and Dian Jiao and Jinbao Xue and Xipeng Zhang and Decheng Wu and Kai Liu and Dengpeng Wu and Guanghui Xu and Shaohua Chen and Shuang Chen and Xiao Feng and Yigeng Hong and Junqiang Zheng and Chengcheng Xu and Zongwei Li and Xiong Kuang and Jianglu Hu and Yiqi Chen and Yuchi Deng and Guiyang Li and Ao Liu and Chenchen Zhang and Shihui Hu and Zilong Zhao and Zifan Wu and Yao Ding and Weichao Wang and Han Liu and Roberts Wang and Hao Fei and Peijie She and Ze Zhao and Xun Cao and Hai Wang and Fusheng Xiang and Mengyuan Huang and Zhiyuan Xiong and Bin Hu and Xuebin Hou and Lei Jiang and Jiajia Wu and Yaping Deng and Yi Shen and Qian Wang and Weijie Liu and Jie Liu and Meng Chen and Liang Dong and Weiwen Jia and Hu Chen and Feifei Liu and Rui Yuan and Huilin Xu and Zhenxiang Yan and Tengfei Cao and Zhichao Hu and Xinhua Feng and Dong Du and Tinghao She and Yangyu Tao and Feng Zhang and Jianchen Zhu and Chengzhong Xu and Xirui Li and Chong Zha and Wen Ouyang and Yinben Xia and Xiang Li and Zekun He and Rongpeng Chen and Jiawei Song and Ruibin Chen and Fan Jiang and Chongqing Zhao and Bo Wang and Hao Gong and Rong Gan and Winston Hu and Zhanhui Kang and Yong Yang and Yuhong Liu and Di Wang and Jie Jiang},
      year={2024},
      journal = {CoRR},
      volume = {abs/2411.02265},
      url={https://arxiv.org/abs/2411.02265}, 
}
\clearpage
\appendix

\section{Efficiency Analyses}
\label{apdx:efficiency}

\begin{figure}[]
    \centering
    \includegraphics[width=\linewidth]{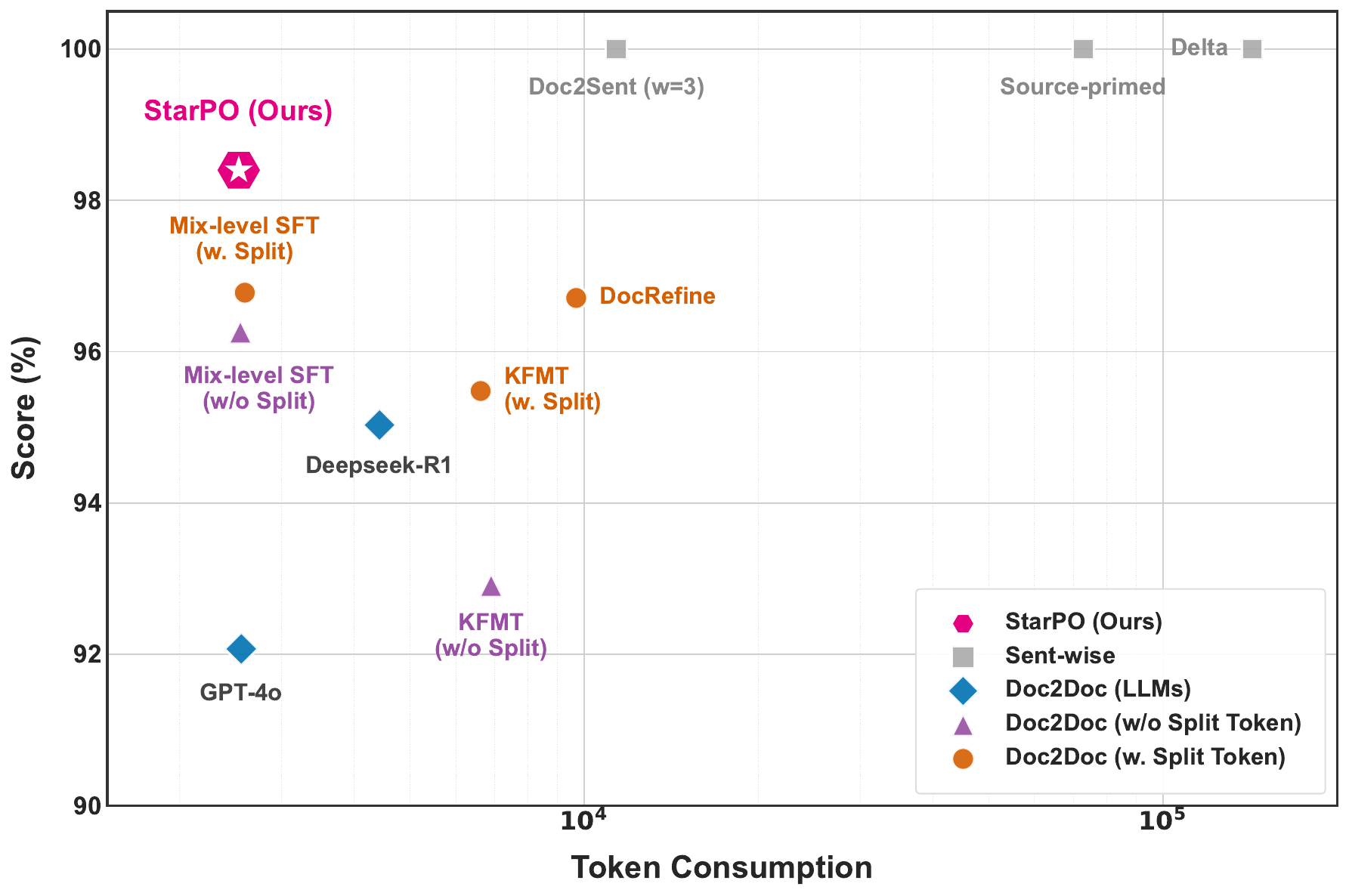}
    \caption{\textbf{Alignment score vs. Token Consumption.} 
    We compare our proposed \textbf{StarPO} against various baselines, including sentence-level systems, document-level baselines, and proprietary LLMs (e.g., GPT-4o, Deepseek-R1). 
    Note that the x-axis is plotted on a logarithmic scale ($10^n$).}
    \label{fig:token_efficiency}
\end{figure}

We evaluate our framework against several representative document translation paradigms. Figure~\ref{fig:token_efficiency} compares the alignment score vs. token consumption among several typical Doc2Sent and Doc2Doc systems. Doc2Sent (w=3)~\cite{wu-etal-2024-adapting,lyu-etal-2024-dempt,koneru-etal-2024-contextual,cui-etal-2024-efficiently} utilizes a standard sliding window of three preceding sentences for context, while Doc2Sent (Source-primed)~\cite{hu-etal-2025-source} treats translation as a sequential multi-turn dialogue. Doc2Sent (DelTA)~\cite{wang-etal-2025-delta} employs an agentic framework for autonomous context management. Among document-level models, Doc2Doc (KFMT)~\cite{liu-etal-2025-improving} and Doc2Doc (Mix-level SFT)~\cite{li-etal-2026-enhancing} use multi-turn interactions or specialized fine-tuning, evaluated both with and without sentence delimiters. Doc2Doc (DocRefine)~\cite{dong-etal-2025-two} focuses on iterative multi-stage refinement to improve coherence. For large-scale baselines, we include Doc2Doc (DeepSeek-R1) to represent reasoning-based models and Doc2Doc (GPT-4o) as a high-performance proprietary benchmark. Finally, Doc2Doc (StarPO) is our proposed framework, which uses STAR-curated preference pairs to optimize for both structural alignment and semantic fidelity.

\section{Detailed Case Study}
\label{apdx:detail_case_study}
\begin{figure*}[]
	\centering
	\includegraphics[width=\linewidth]{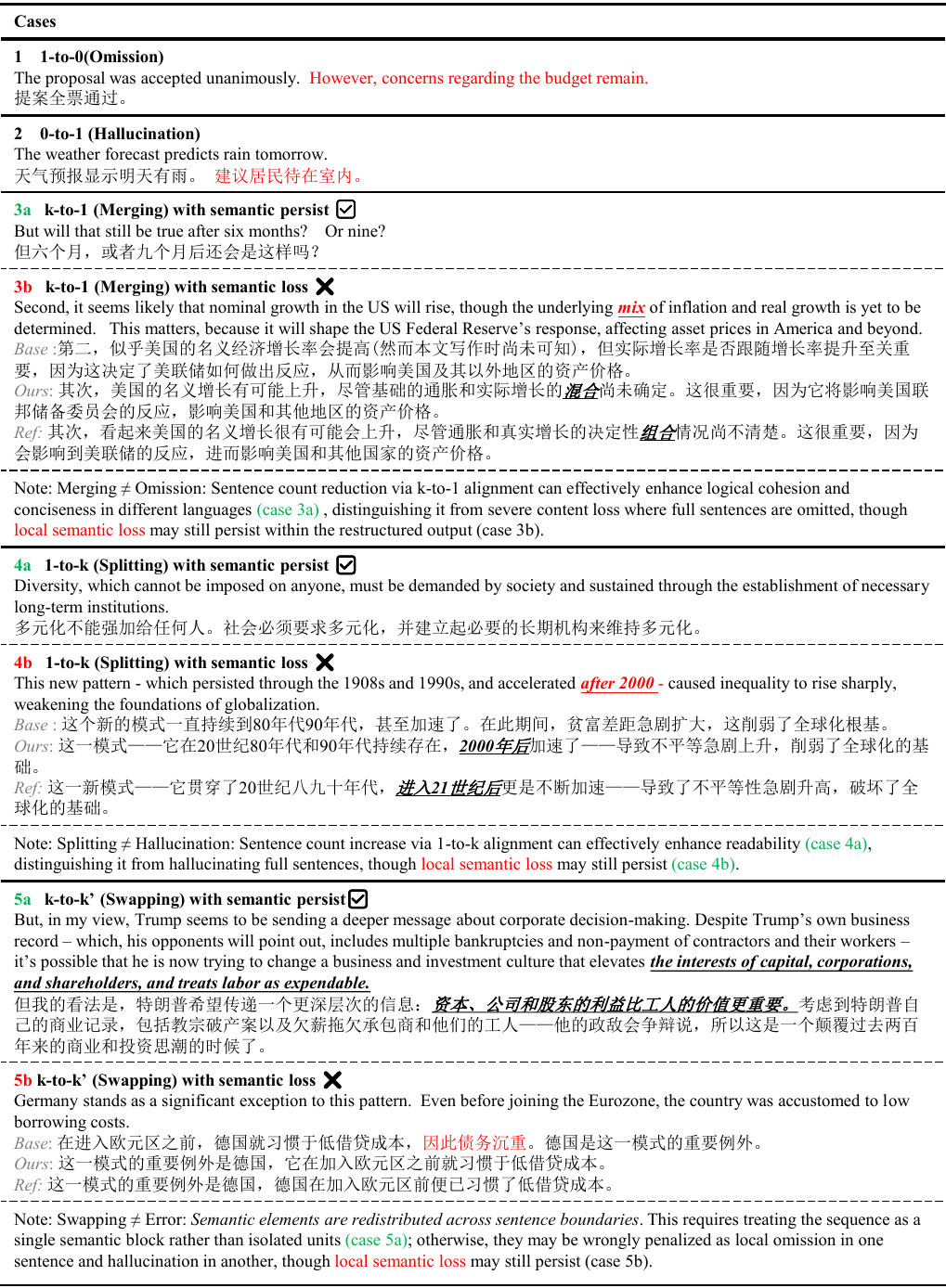}
    \caption{Case study of structural alignment patterns. \textcolor{MyGreen}{Green text} indicates acceptable linguistic adaptations (e.g., merging for cohesion or splitting for clarity), while \textcolor{red}{red text} highlights genuine translation errors. These two scenarios must be carefully distinguished to avoid misjudging stylistic alignment as semantic loss.}	\label{fig:case_study}
\end{figure*}
\label{apdx:detailed_case_study}

While STAR prioritizes 1-to-1 alignment, Figure~\ref{fig:case_study} illustrates that structural deviations range from pathological errors to valid linguistic adaptations. We categorize these patterns based on their semantic impact:

\begin{itemize}
\item \textbf{Pathological Deviations:} 1-to-0 (Omission) and 0-to-1 (Hallucination) instances that result in clear fidelity loss (\textcolor{red}{red}).
\item \textbf{Structural Adaptations:} $k$-to-1 (Merging), 1-to-$k$ (Splitting), and $k$-to-$k'$ (Swapping) patterns that often optimize cohesion or readability (\textcolor{MyGreen}{green}).
\end{itemize}

As detailed in the figure's notes, even valid adaptations (Cases 3a, 4a, 5a) may harbor localized semantic inaccuracies (Cases 3b, 4b, 5b), necessitating a nuanced evaluation that distinguishes stylistic alignment from actual information loss.

\section{Impact of Input Context Length on Structural Alignment}
\label{apdx:length_vs_star}

\begin{figure}[h]
    \centering
    \includegraphics[width=\linewidth]{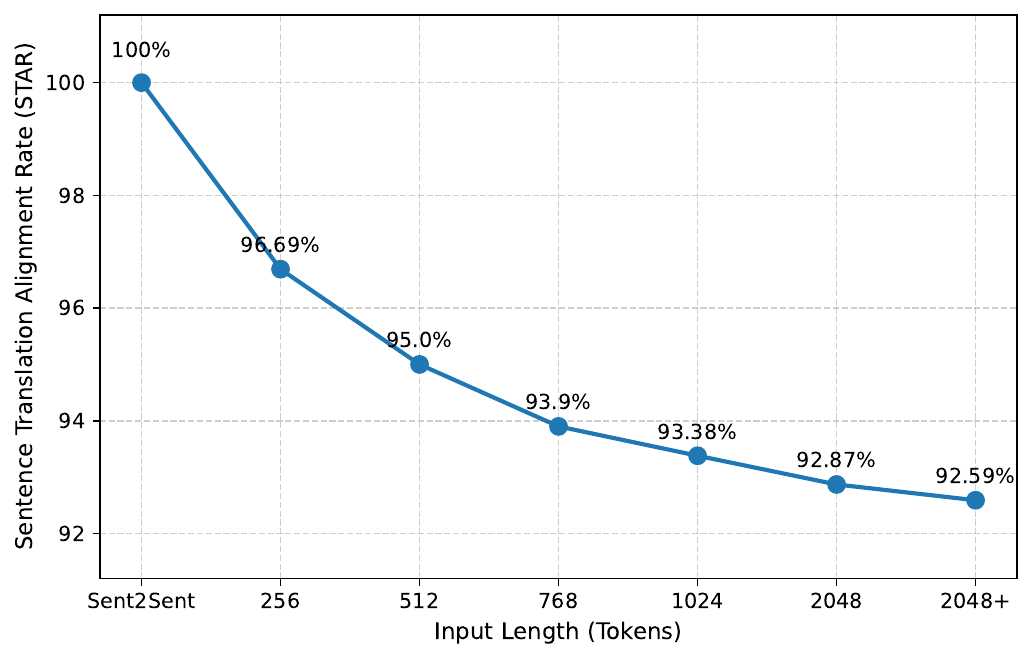}
   \caption{Sentence Translation Alignment Rate (STAR) across varying input context lengths from Sent2Sent to more than 2048 tokens. The results are evaluated on Zh $\Rightarrow$ En translation of the News Commentary dataset using LLaMA-3.1-8B-Instruct.}
    \label{fig:len_star}
\end{figure}

To further investigate the relationship between document length and structural fidelity in Doc2Doc translation, we evaluated the Sentence Translation Alignment Rate (STAR) across varying input context lengths using LLaMA-3-8B-Instruct on News-Commentary test set on Zh $\Rightarrow$ En translation. As illustrated in Figure~\ref{fig:len_star}, there is a clear negative correlation between the input length (measured in tokens) and the alignment quality. 

In the standard sentence-to-sentence (Sent2Sent) translation paradigm, the model naturally achieves a perfect 100\% STAR score. However, as the input context expands to 256 tokens, the alignment rate sharply drops to 96.69\%. This highlights the critical need for explicit structural alignment objectives, such as our proposed StarPO, particularly for long-document translation.

\section{LLM-judge version of STAR}
\label{apdx:llm_judge_of_star}
\begin{figure*}[t]
	\centering
	\includegraphics[width=\linewidth]{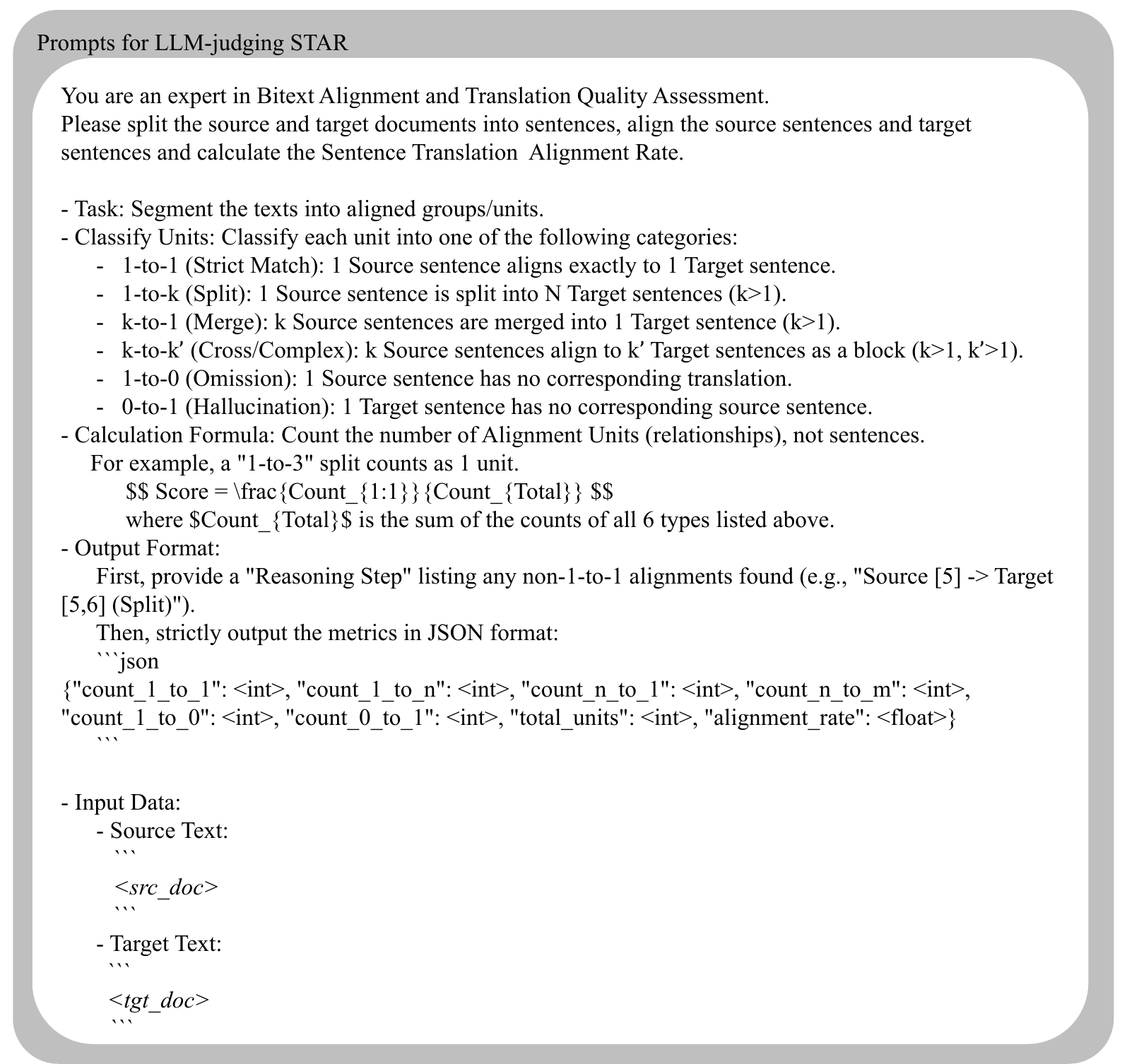}
    \caption{The prompt template used for the LLM-as-a-judge implementation of STAR. The final score is calculated strictly based on the ratio of 1-to-1 matches to total alignment units, serving as a high-precision reference for validating our automated metric.} 
\label{fig:llm_star_prompt}
\end{figure*}

To validate the accuracy of our automated STAR metric, we implement an LLM-based version using the prompt template illustrated in Figure~\ref{fig:llm_star_prompt}. This prompt is designed to simulate human-level judgment on document structure.

\section{Discussion of Threshold $\tau$}
\label{apdx:tau}

To investigate the impact of the margin threshold $\tau$ used in preference data construction, we conduct a sensitivity analysis across both Zh $\Rightarrow$ En and En $\Rightarrow$ Zh directions. As shown in Table~\ref{tab:tau}, the performance variation remains minimal (less than 0.6 points) as $\tau$ ranges from $0.00$ to $0.15$. This stability suggests that StarPO is robust to the choice of $\tau$ and does not require extensive hyperparameter tuning. We select $\tau = 0.10$ as the default for all experiments, as it yields a marginal but consistent advantage in our pilot studies.

\begin{table}[]
\centering
 \small
	\begin{NiceTabular}{lcc}
 \toprule
		$\bf \tau$ & \bf Zh $\Rightarrow$ En & \bf En $\Rightarrow$ Zh \\
  \midrule
 0.00 & 81.04 & 80.01   \\
 0.05  & 81.15 & 79.96  \\
 0.10 & 81.55 & 80.11 \\
 0.15 & 81.38 & 80.08 \\
 \midrule
	\end{NiceTabular}
	\caption{Effect of $\tau$ on Zh $\Leftrightarrow$ En translations.}
	\label{tab:tau}
\end{table}

\section{Data Statistics}
\label{apdx:dataset stat}
In this section, we provide detailed statistics regarding the bilingual parallel corpora utilized in our experiments: the News-Commentary dataset and the Guofeng dataset. The statistical summaries for these datasets are presented in Table~\ref{table:news_dataset} and Table~\ref{table:guofeng_dataset}, respectively.

The News-Commentary dataset encompasses a diverse range of language pairs, including Chinese-English (ZH$\Leftrightarrow$EN), Chinese-German (ZH$\Leftrightarrow$DE), Russian-English (RU$\Leftrightarrow$EN), German-English (DE$\Leftrightarrow$EN), and Spanish-English (ES$\Leftrightarrow$EN). The Guofeng dataset primarily focuses on pairs involving Chinese, specifically Chinese-Russian (ZH$\Leftrightarrow$RU), Chinese-English (ZH$\Leftrightarrow$EN), and Chinese-German (ZH$\Leftrightarrow$DE).

For each dataset and language direction, we report four key metrics based on document-level analysis: the quantity of documents across the training, validation, and test splits; the size of the dataset used for preference optimization; the average number of tokens per document; and the maximum number of tokens found in a single document within the corpus.

\begin{table}[t]
\centering
\small
\setlength{\tabcolsep}{2pt}
\begin{NiceTabular}{c|cccc}
\toprule
\textbf{Dataset}& \begin{tabular}[c]{@{}c@{}}\textbf{\#Document}\\ \textbf{Train/Valid/Test}\end{tabular} & \begin{tabular}[c]{@{}c@{}}\textbf{\#Document}\\ \textbf{for StarPO}\end{tabular} & \begin{tabular}[c]{@{}c@{}}\textbf{Average}\\ \textbf{Tokens}\end{tabular} & \begin{tabular}[c]{@{}c@{}}\textbf{Max}\\ \textbf{Tokens}\end{tabular} \\ 
\midrule
De $\Rightarrow$ En & 8.4K/150/150 & 1,008 & 1,797 & 6,540 \\ 

En $\Rightarrow$ De & 8.4K/150/150 & 734 & 1,066 & 4,065 \\ 
\midrule
Es $\Rightarrow$ En & 9.7K/150/150 & 3,325 & 1,643 & 6,293 \\ 
En $\Rightarrow$ Es & 9.7K/150/150 & 3,937 & 1,071 & 4,146 \\ 
\midrule
Ru $\Rightarrow$ En & 7.3K/150/150 & 1,744  & 1,776 & 7,951 \\ 
En $\Rightarrow$ Ru & 7.3K/150/150 & 1,491 & 1,079 & 5,557 \\ 

\midrule
Zh $\Rightarrow$ En & 8.6K/150/150 & 1,253 & 1,377 & 4,425 \\ 
En $\Rightarrow$ Zh & 8.6K/150/150 & 1,753 & 1,090 & 3,609 \\ 

\midrule
Zh $\Rightarrow$ De & 7.7K/150/150 & 2,504 & 1,357 & 5,912 \\ 
De $\Rightarrow$ Zh & 7.7K/150/150 & 1,364 & 1,828 & 7,215 \\ 

\bottomrule
\end{NiceTabular}
\caption{Statistics of the News-Commentary dataset.}
\label{table:news_dataset}
\end{table}

\begin{table}[t]
\centering
\small
\setlength{\tabcolsep}{2pt}
\begin{NiceTabular}{c|cccc}
\toprule
\textbf{Dataset}& \begin{tabular}[c]{@{}c@{}}\textbf{\#Document}\\ \textbf{Train/Valid/Test}\end{tabular} & \begin{tabular}[c]{@{}c@{}}\textbf{\#Document}\\ \textbf{for StarPO}\end{tabular} & \begin{tabular}[c]{@{}c@{}}\textbf{Average}\\ \textbf{Tokens}\end{tabular} & \begin{tabular}[c]{@{}c@{}}\textbf{Max}\\ \textbf{Tokens}\end{tabular} \\ 
\midrule
Zh $\Rightarrow$ En & 22.0K/25/25 & 2.0K & 1,853 & 12,961 \\ 
En $\Rightarrow$ Zh & 22.0K/25/25 & 2.0K & 1,624 &  10,956 \\ 
\midrule
Zh $\Rightarrow$ De & 6.0K/30/30 & 2.0K &  1,927 & 7,962 \\ 
De $\Rightarrow$ Zh & 6.0K/30/30 & 2.0K & 2,392 & 9,909 \\ 

\midrule
Zh $\Rightarrow$ Ru & 6.0K/30/30 & 2.0K & 1,991 & 6,134\\ 
Ru $\Rightarrow$ Zh & 6.0K/30/30 & 2.0K & 2,470 & 7,514\\ 
\bottomrule
\end{NiceTabular}
\caption{Statistics of the Guofeng dataset.}
\label{table:guofeng_dataset}
\end{table}

\section{Implementation Details}
\label{apdx:implementation_details}
\begin{figure}[]
	\centering
	\includegraphics[width=\columnwidth]{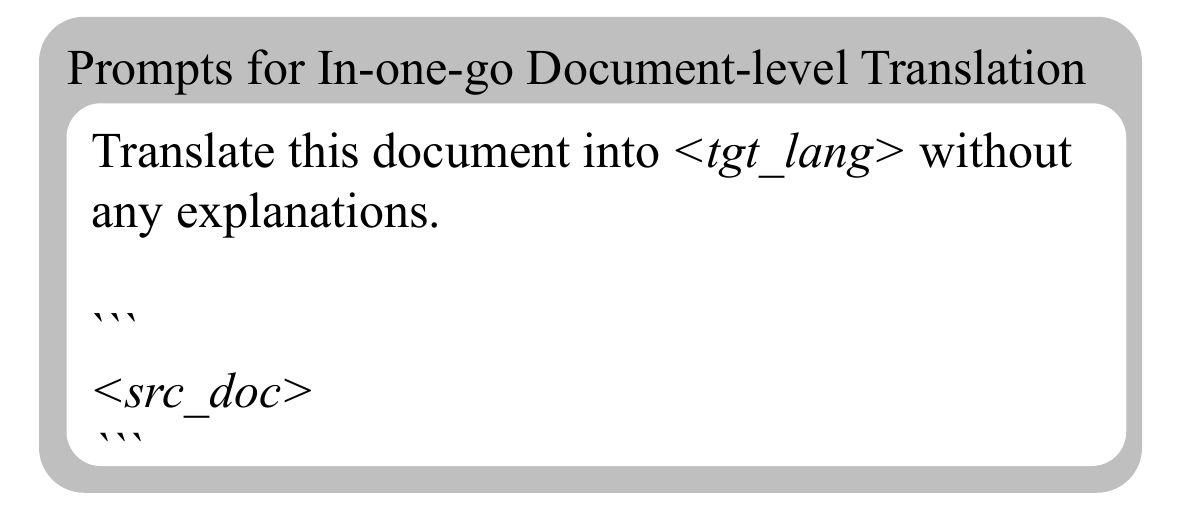}
	\caption{\textbf{The universal prompt template used for document-level translation.} To ensure a fair comparison and eliminate prompt engineering variance, we apply this standardized Doc2Doc instruction across all experiments. \textit{<tgt\_lang>} represents the target language, and \textit{<src\_doc>} represents the source document, respectively.}
    \label{fig:doc2doc_prompt}
\end{figure}

We implement our experiments in ms-swift~\cite{zhao-etal-2025-swift}\footnote{\url{https://github.com/modelscope/ms-swift}}. During fine-tuning, we adapt LoRA~\cite{hu-etal-2021-lora}. We set LoRA rank to 8 and LoRA alpha to 16 , respectively. The models are trained for 1 epoch using AdamW optimizer with learning rate of $1\times 10^{-4}$, warmup ratio of 0.05. We set $\beta$ to 0.1. Our experiments run on one NVIDIA H100 GPU, requiring approximately 1 hour for training. Regarding evaluation efficiency, our proposed STAR metric is computationally efficient, achieving a throughput of approximately 3 samples per second. During inferencing, we set temperature to 0.3, beam size to 1 in vllm\footnote{\url{https://github.com/vllm-project/vllm}} framework. The specific prompt template used in our experiments is illustrated in Figure~\ref{fig:doc2doc_prompt}.

\section{STAR Scores}
\label{apdx:Detailed_metric_scores}

\begin{table*}[]
\centering
\small
\renewcommand{\arraystretch}{1.2} 

\newcommand{\mystack}[2]{%
  \begin{tabular}{@{}c@{}}
    #1 \\[-0.5ex] 
    \footnotesize\color{gray}#2
  \end{tabular}%
}

\begin{NiceTabular}{lccccccccccc}
\toprule
\Block[l]{2-1}{\textbf{System}} & \Block[c]{1-2}{\textbf{Zh $\Leftrightarrow$ En}} & & \Block[c]{1-2}{\textbf{De $\Leftrightarrow$ En}} & & \Block[c]{1-2}{\textbf{De $\Leftrightarrow$ Zh}} & & \Block[c]{1-2}{\textbf{Ru $\Leftrightarrow$ En}} & & \Block[c]{1-2}{\textbf{En $\Leftrightarrow$ Es}} & & \Block[l]{2-1}{\textbf{Avg.}}\\
\cmidrule(lr){2-3}\cmidrule(lr){4-5}\cmidrule(lr){6-7}\cmidrule(lr){8-9}\cmidrule(lr){10-11}
& $\Rightarrow$ & $\Leftarrow$ & $\Rightarrow$ & $\Leftarrow$ & $\Rightarrow$ & $\Leftarrow$& $\Rightarrow$ & $\Leftarrow$& $\Rightarrow$ & $\Leftarrow$ \\ 
\midrule

\rowcolor[gray]{0.9}
\Block[c]{1-12}{\textsc{LLaMA-3.1-8B-Instruct}} \\    
Base      & \mystack{90.31}{96.44} & \mystack{87.21}{93.65} & \mystack{92.52}{94.43} & \mystack{95.41}{95.92} & \mystack{78.96}{93.55} & \mystack{82.79}{89.37} & \mystack{97.05}{98.51} & \mystack{93.92}{96.46} & \mystack{90.97}{96.98} & \mystack{94.06}{95.86} & \mystack{90.32}{95.12} \\
+ SFT     & \mystack{93.52}{95.00} & \mystack{91.75}{95.05} & \mystack{95.77}{96.39} & \mystack{95.60}{97.03} & \mystack{80.39}{94.38} & \mystack{93.58}{96.28} & \mystack{96.97}{98.58} & \mystack{97.58}{97.82} & \mystack{77.13}{96.04} & \mystack{95.64}{96.27} & \mystack{91.79}{96.28} \\
\hspace{1em}+ CPO     & \mystack{95.00}{97.05} & \mystack{91.72}{96.62} & \mystack{95.99}{96.90} & \mystack{95.52}{\underline{98.76}} & \mystack{79.83}{97.93} & \mystack{90.41}{97.42} & \mystack{97.30}{98.91} & \mystack{97.65}{97.88} & \mystack{90.64}{91.87} & \mystack{97.50}{97.82} & \mystack{93.16}{97.12} \\
\hspace{1em}+StarPO   & \mystack{\underline{95.62}}{98.64} & \mystack{93.36}{98.31} & \mystack{\underline{96.80}}{97.08} & \mystack{95.59}{98.66} & \mystack{80.80}{98.36} & \mystack{\textbf{97.12}}{97.54} & \mystack{97.27}{98.90} & \mystack{\textbf{98.01}}{98.08} & \mystack{90.13}{92.45} & \mystack{\textbf{97.63}}{98.01} & \mystack{94.23}{97.60} \\ \midrule

\rowcolor[gray]{0.9}
\Block[c]{1-12}{\textsc{Qwen2.5-7B-Instruct}} \\ 
Base      & \mystack{94.97}{96.42} & \mystack{93.92}{95.72} & \mystack{95.85}{96.72} & \mystack{95.10}{96.11} & \mystack{61.80}{72.68} & \mystack{92.03}{94.38} & \mystack{\underline{98.05}}{98.46} & \mystack{86.87}{92.16} & \mystack{96.02}{96.77} & \mystack{97.33}{97.76} & \mystack{91.19}{93.72} \\
+SFT      & \mystack{95.51}{96.90} & \mystack{92.96}{98.85} & \mystack{94.54}{96.88} & \mystack{94.45}{96.66} & \mystack{88.42}{92.00} & \mystack{93.25}{97.20} & \mystack{97.49}{98.82} & \mystack{95.55}{98.99} & \mystack{81.10}{95.38} & \mystack{96.80}{97.32} & \mystack{93.01}{96.90} \\
\hspace{1em}+CPO      & \mystack{95.40}{98.93} & \mystack{91.09}{\textbf{99.00}} & \mystack{95.89}{98.96} & \mystack{95.36}{96.63} & \mystack{95.92}{96.04} & \mystack{92.49}{97.12} & \mystack{97.64}{99.08} & \mystack{95.54}{99.32} & \mystack{92.83}{92.94} & \mystack{\underline{97.56}}{99.15} & \mystack{94.97}{97.72} \\
\hspace{1em}+StarPO    & \mystack{\textbf{96.07}}{\textbf{99.29}} & \mystack{\underline{95.67}}{98.77} & \mystack{96.55}{\underline{99.45}} & \mystack{95.36}{96.71} & \mystack{\textbf{96.83}}{\textbf{99.01}} & \mystack{93.12}{97.35} & \mystack{\textbf{98.15}}{\textbf{99.68}} & \mystack{\underline{97.71}}{\textbf{99.74}} & \mystack{\underline{96.21}}{97.01} & \mystack{97.37}{99.60} & \mystack{\textbf{96.30}}{\underline{98.66}} \\ \midrule

\rowcolor[gray]{0.9}
\Block[c]{1-12}{\textsc{Qwen3-4B-Instruct}} \\ 
Base      & \mystack{94.67}{98.73} & \mystack{93.41}{94.02} & \mystack{96.33}{98.08} & \mystack{\underline{96.45}}{96.81} & \mystack{82.66}{93.10} & \mystack{91.01}{94.79} & \mystack{97.69}{98.08} & \mystack{96.92}{97.12} & \mystack{96.12}{96.21} & \mystack{91.38}{94.95} & \mystack{93.66}{96.19} \\
+SFT      & \mystack{95.02}{98.90} & \mystack{92.92}{98.91} & \mystack{96.48}{99.11} & \mystack{91.05}{98.48} & \mystack{83.03}{96.44} & \mystack{91.26}{96.76} & \mystack{97.48}{98.97} & \mystack{95.35}{99.63} & \mystack{95.36}{97.14} & \mystack{96.12}{99.58} & \mystack{93.41}{98.39} \\
\hspace{1em}+CPO      & \mystack{95.01}{98.93} & \mystack{93.24}{98.90} & \mystack{96.60}{\textbf{100.0}} & \mystack{95.33}{\textbf{98.99}} & \mystack{92.16}{96.74} & \mystack{92.30}{97.57} & \mystack{97.63}{99.05} & \mystack{97.57}{97.91} & \mystack{92.13}{96.58} & \mystack{97.16}{\textbf{99.71}} & \mystack{94.91}{98.44} \\
\hspace{1em}+StarPO    & \mystack{\textbf{96.07}}{\underline{99.13}} & \mystack{95.65}{\underline{98.99}} & \mystack{96.55}{\textbf{100.0}} & \mystack{95.36}{98.62} & \mystack{95.85}{\underline{98.45}} & \mystack{\underline{95.85}}{\textbf{98.92}} & \mystack{97.88}{\underline{99.35}} & \mystack{97.58}{\underline{99.70}} & \mystack{93.80}{\textbf{98.89}} & \mystack{96.45}{\underline{99.62}} & \mystack{\underline{96.10}}{\textbf{99.17}} \\ \midrule

\rowcolor[gray]{0.9}
\Block[c]{1-12}{\textsc{Other Systems}}   \\
Tower+    & \mystack{93.02}{94.15} & \mystack{94.89}{95.45} & \mystack{93.13}{95.79} & \mystack{95.50}{96.50} & \mystack{94.93}{96.03} & \mystack{93.78}{96.02} & \mystack{96.46}{98.04} & \mystack{96.04}{98.04} & \mystack{\textbf{96.37}}{\underline{98.16}} & \mystack{95.43}{97.78} & \mystack{94.96}{96.60} \\
GPT-4o    & \mystack{93.07}{94.66} & \mystack{91.28}{96.76} & \mystack{91.58}{97.38} & \mystack{92.10}{97.26} & \mystack{94.82}{98.23} & \mystack{92.86}{96.39} & \mystack{93.74}{97.53} & \mystack{93.76}{97.46} & \mystack{93.88}{97.54} & \mystack{93.14}{96.80} & \mystack{93.02}{97.00} \\
Deepseek-R1 & \mystack{93.56}{94.04} & \mystack{\textbf{96.51}}{97.81} & \mystack{\textbf{97.33}}{97.44} & \mystack{\textbf{97.38}}{97.88} & \mystack{\underline{96.22}}{96.26} & \mystack{94.12}{\underline{98.03}} & \mystack{90.39}{92.49} & \mystack{93.25}{97.05} & \mystack{93.08}{97.22} & \mystack{97.09}{97.12} & \mystack{94.89}{96.53} \\
\bottomrule
\end{NiceTabular}
\caption{Performance in STAR scores on News-Commentary test set. In each cell, the top value is the \textbf{Strict} score and the bottom value (in gray) is the \textbf{Relaxed} score.}
\label{table:STAR on news}
\end{table*}

\begin{table}[]
\centering
\small
\renewcommand{\arraystretch}{1.2} 
\setlength{\tabcolsep}{3pt}    

\newcommand{\mystack}[2]{%
  \begin{tabular}{@{}c@{}}
    #1 \\[-0.5ex] 
    \footnotesize\color{gray}#2
  \end{tabular}%
}

\resizebox{\columnwidth}{!}{
\begin{NiceTabular}{lcccccc}
\toprule
\Block[l]{2-1}{\textbf{System}} & \Block[c]{1-2}{\textbf{Zh $\Leftrightarrow$ En}} & & \Block[c]{1-2}{\textbf{Zh $\Leftrightarrow$ De}} & & \Block[c]{1-2}{\textbf{Zh $\Leftrightarrow$ Ru}}\\
\cmidrule(lr){2-3}\cmidrule(lr){4-5}\cmidrule(lr){6-7}
& $\Rightarrow$ & $\Leftarrow$ & $\Rightarrow$ & $\Leftarrow$ & $\Rightarrow$ & $\Leftarrow$ \\ 
\midrule
\rowcolor[gray]{0.9}
\Block[c]{1-7}{\textsc{LLaMA-3.1-Instruct}}                                     \\ 
Base      & \mystack{30.12}{38.84} & \mystack{57.60}{72.33} & \mystack{16.40}{23.30} & \mystack{28.65}{35.57} & \mystack{18.37}{35.53} & \mystack{31.70}{44.35} \\
+ SFT     & \mystack{70.94}{80.51} & \mystack{61.46}{83.67} & \mystack{42.68}{78.67} & \mystack{60.62}{69.59} & \mystack{78.63}{80.89} & \mystack{50.25}{62.60} \\
\hspace{0.5em}+ CPO     & \mystack{72.80}{80.74} & \mystack{74.15}{77.01} & \mystack{73.41}{80.06} & \mystack{80.96}{85.98} & \mystack{75.81}{80.03} & \mystack{87.34}{98.58} \\
\hspace{0.5em}+ StarPO  & \mystack{74.29}{82.37} & \mystack{76.84}{83.05} & \mystack{75.97}{79.39} & \mystack{\underline{84.39}}{88.32} & \mystack{80.34}{85.40} & \mystack{\underline{87.94}}{97.39} \\ \midrule
\rowcolor[gray]{0.9}
\Block[c]{1-7}{\textsc{Qwen2.5-7B-Instruct}}  \\ 
Base      & \mystack{60.30}{95.83} & \mystack{75.01}{91.61} & \mystack{69.90}{93.56} & \mystack{42.02}{91.93} & \mystack{64.10}{92.94} & \mystack{82.58}{91.63} \\
+SFT      & \mystack{58.42}{94.71} & \mystack{90.47}{93.79} & \mystack{75.36}{95.44} & \mystack{61.71}{88.59} & \mystack{63.39}{78.87} & \mystack{80.81}{93.93} \\
\hspace{0.5em}+CPO      & \mystack{71.16}{97.13} & \mystack{89.04}{94.39} & \mystack{72.75}{90.66} & \mystack{75.09}{93.50} & \mystack{\underline{81.81}}{88.60} & \mystack{81.39}{93.91} \\
\hspace{0.5em}+StarPO   & \mystack{74.64}{96.91} & \mystack{91.48}{92.93} & \mystack{\textbf{78.07}}{93.79} & \mystack{86.23}{95.70} & \mystack{\textbf{82.21}}{95.06} & \mystack{87.47}{94.25} \\ \midrule
\rowcolor[gray]{0.9}
\Block[c]{1-7}{\textsc{Qwen3-4B-Instruct}} \\ 
Base      & \mystack{54.98}{95.62} & \mystack{76.23}{77.27} & \mystack{28.15}{94.85} & \mystack{84.42}{87.47} & \mystack{69.53}{89.53} & \mystack{55.95}{91.06} \\
+SFT      & \mystack{77.63}{89.92} & \mystack{89.55}{92.07} & \mystack{43.49}{85.10} & \mystack{81.99}{84.73} & \mystack{71.58}{91.98} & \mystack{81.56}{97.63} \\
\hspace{0.5em}+CPO      & \mystack{\underline{78.29}}{89.87} & \mystack{\textbf{90.11}}{92.48} & \mystack{67.10}{87.67} & \mystack{87.16}{87.45} & \mystack{72.77}{82.37} & \mystack{76.54}{85.71} \\
\hspace{0.5em}+StarPO   & \mystack{\textbf{83.74}}{89.82} & \mystack{\underline{89.65}}{93.11} & \mystack{\underline{77.04}}{95.45} & \mystack{87.47}{88.31} & \mystack{76.87}{87.04} & \mystack{83.18}{97.20} \\ \midrule
\rowcolor[gray]{0.9}
\Block[c]{1-7}{\textsc{Other Systems}} \\ 
Tower+    & \mystack{62.68}{90.83} & \mystack{84.39}{93.52} & \mystack{72.87}{90.37} & \mystack{\textbf{86.46}}{94.01} & \mystack{69.10}{87.47} & \mystack{\textbf{91.44}}{96.21} \\
GPT-4o    & \mystack{65.09}{90.06} & \mystack{82.74}{90.04} & \mystack{70.31}{80.81} & \mystack{72.73}{87.23} & \mystack{77.58}{89.84} & \mystack{66.88}{93.29} \\ 
Deepseek  & \mystack{64.00}{89.87} & \mystack{66.57}{82.72} & \mystack{69.84}{90.24} & \mystack{64.52}{93.30} & \mystack{67.78}{92.74} & \mystack{62.72}{90.84} \\
\bottomrule
\end{NiceTabular}
}
\caption{STAR scores on Guofeng test set. In each cell, the top value is the \textbf{Strict} score and the bottom value (in gray) is the \textbf{Relaxed} score.}
\label{table:STAR on Guofeng testset}
\end{table}

\begin{table}[]
\centering
\small
\begin{NiceTabular}{lcccc}
\toprule
\Block{2-1}{\textbf{System}} & \textbf{Ideal} & \multicolumn{3}{c}{\textbf{Structural Deviations}} \\
\cmidrule(lr){2-2} \cmidrule(lr){3-5}
 & \textbf{1-to-1} & \textbf{1-to-0} & \textbf{0-to-1} & \textbf{Other} \\
\midrule

\rowcolor[gray]{0.9}
\Block[c]{1-5}{\textsc{LLaMA-3.1-8B-Instruct}} \\
Base & 92.59 & 2.08 & 1.17 & 4.16 \\
+SFT & 93.73 & 2.72 & 0.42 & 3.13 \\
\hspace{0.5em}+CPO & 95.42 & 0.33 & 0.75 & 3.50 \\
\hspace{0.5em}+StarPO & 95.79 & 1.95 & 0.14 & 2.12 \\
\midrule

\rowcolor[gray]{0.9}
\Block[c]{1-5}{\textsc{Qwen-2.5-7B-Instruct}} \\
Base & 95.35 & 1.91 & 1.31 & 1.43 \\
+SFT & 96.63 & 0.98 & 1.60 & 0.79 \\
\hspace{0.5em}+CPO & 97.46 & 0.35 & 1.52 & 0.67 \\
\hspace{0.5em}+StarPO & 98.43 & 0.68 & 0.00 & 0.89 \\
\midrule

\rowcolor[gray]{0.9}
\Block[c]{1-5}{\textsc{Qwen3-4B-Instruct}} \\
Base & 94.72 & 0.72 & 0.33 & 4.23 \\
+SFT & 95.36 & 0.80 & 0.00 & 3.84 \\
\hspace{0.5em}+CPO & 98.02 & 0.22 & 0.00 & 1.76 \\
\hspace{0.5em}+StarPO & 98.09 & 0.64 & 0.00 & 1.27 \\
\midrule

\rowcolor[gray]{0.9}
\Block[c]{1-5}{\textsc{Other Systems}} \\
Tower+ & 94.48 & 2.68 & 0.74 & 2.10 \\
GPT-4o & 92.91 & 2.25 & 2.89 & 1.95 \\
Deepseek-R1 & 95.03 & 4.85 & 0.03 & 0.09 \\
\bottomrule
\end{NiceTabular}
\caption{Calculating STAR scores in Chinese $\Rightarrow$ English language direction on News-Commentary test set by Gemini-2.5-Flash using prompts in Figure ~\ref{fig:llm_star_prompt}.}
\label{tab:full_alignment_stats}
\end{table}

The structural fidelity of these systems, as measured by our proposed STAR score, is detailed in Tables \ref{table:STAR on news} and \ref{table:STAR on Guofeng testset}. Additionally, we report the alignment results on the News-Commentary Zh $\Rightarrow$ En dataset evaluated by Gemini-2.5-Flash using the prompt from Appendix~\ref{apdx:llm_judge_of_star}, as shown in Table~\ref{tab:full_alignment_stats}.

\section{Detailed BLEU Scores on Guofeng Dataset}
\label{apdx:guofeng_bleu}
Table~\ref{table:BLEU on Guofeng testset} shows detailed BLEU Scores on Guofeng Dataset.
\begin{table}[]
\centering
\small
\resizebox{\columnwidth}{!}{
\begin{NiceTabular}{lcccccc}
\toprule
\Block[l]{2-1}{\textbf{System}} & \Block[c]{1-2}{\textbf{Zh $\Leftrightarrow$ En}} & & \Block[c]{1-2}{\textbf{Zh $\Leftrightarrow$ De}} & & \Block[c]{1-2}{\textbf{Zh $\Leftrightarrow$ Ru}}\\
\cmidrule(lr){2-3}\cmidrule(lr){4-5}\cmidrule(lr){6-7}
& $\Rightarrow$ & $\Leftarrow$ & $\Rightarrow$ & $\Leftarrow$ & $\Rightarrow$ & $\Leftarrow$ \\ 
\midrule
\rowcolor[gray]{0.9}
\Block[c]{1-7}{\textsc{Llama-3.1-8B-Instruct}}                                                   \\ 
Base & 8.36 & 16.56 & 4.41 & 6.13 & 13.57 & 12.68 \\
+ SFT & 9.41 & 17.45 & 14.83 & 15.55 & 19.78 & 14.09  \\
\hspace{0.5em}+ CPO  & 11.25 & 19.19 & 17.24 & 16.13 & 21.67 & 18.45 \\
\hspace{0.5em}+ StarPO & \colorbox{blue!15}{12.08} & 19.60 & \colorbox{blue!15}{18.20} & \colorbox{blue!15}{20.49} & \colorbox{blue!15}{ 23.24} & \colorbox{blue!15}{\underline{19.39}} \\ \midrule
\rowcolor[gray]{0.9}
\Block[c]{1-7}{\textsc{Qwen2.5-7B-Instruct}}                                                  \\ 
Base &18.73 & 18.46 & 14.70 & 18.46 & 14.23 &  9.13   \\
+SFT & 19.19 & 20.35 & 16.99 &  18.98 & 15.72  & 17.19   \\
\hspace{0.5em}+CPO &18.96 &  21.54  & 18.65 & 19.37 & 22.05 & 17.01  \\
\hspace{0.5em}+StarPO & \underline{19.22} &  \colorbox{blue!15}{22.17} & \colorbox{blue!15}{\underline{21.92}} & 20.10 & \colorbox{blue!15}{23.15} &  \colorbox{blue!15}{18.92} \\ \midrule
\rowcolor[gray]{0.9}
\Block[c]{1-7}{\textsc{Qwen3-4B-Instruct}}                                                  \\ 
Base & 17.68 & 17.73 & 12.68 & 14.65 & 11.24 & 15.18 \\
+SFT & 18.37 & 17.95 & 14.67 & 13.72 & 14.78 & 16.50 \\
\hspace{0.5em}+CPO & 17.60 & 19.47 & 18.65 & 19.38 & 21.83 & 17.42 \\
\hspace{0.5em}+StarPO & 18.86 & \colorbox{blue!15}{\underline{23.84}} & \colorbox{blue!15}{19.79} & \colorbox{blue!15}{20.54} & \colorbox{blue!15}{\bf 25.79} & 17.65 \\ \midrule
\rowcolor[gray]{0.9}
\Block[c]{1-7}{\textsc{Other Systems}}  \\ 
Tower+ & \bf 19.74 & 21.23 & \bf 23.21 & \underline{20.98} & 23.34 & 14.98 \\
GPT-4o    & 17.31 & \bf 29.08 & 16.15 & \textbf{27.07} & \underline{25.63} & \bf 27.32 \\ 
Deepseek & 18.57& 19.65 & 14.94& 21.22& 20.38& 16.51\\
\bottomrule
\end{NiceTabular}
}
\caption{Performance in BLEU on Guofeng test set.}
\label{table:BLEU on Guofeng testset}
\end{table}

\section{Details in Preference Optimization}
\label{apdx:offline_po}

We first highlight the practical advantages of STAR over existing model-based metrics. While employing COMET as a reward function for GRPO/GSPO training is computationally expensive, it also exhibits training instability, with the reward loss frequently fluctuating around zero. Furthermore, using COMETKiwi as a reward signal may even lead to off-the-target language generation, where the model erroneously translates the source into an unintended language. In contrast, STAR offers a significantly more efficient and robust alternative, enabling rapid reward calculation (approx. 3 samples/s) while ensuring stable convergence during preference optimization.

To contextualize our method within the broader RLHF landscape, we compare STAR-Masked Preference Optimization against established offline algorithms, including DPO~\cite{rafailov-etal-2023-dpo}, SimPO~\cite{meng-etal-2024-simpo}, KTO~\cite{ethayarajh-etal-2024-kto}, and ORPO~\cite{hong-etal-2024-orpo}. Specifically, we evaluate on the Chinese-to-English subset of the News-Commentary dataset. The results are presented in Table~\ref{table:offline_po_comparison}, indicating that vanilla DPO suffers from severe output collapse; however, adding an SFT loss allows DPO (+SFT) to match the performance of CPO. SimPO also exhibits occasional output collapse on specific entries. While KTO and SimPO prove to be effective, they yield slightly inferior results compared to CPO. Overall, our method demonstrates superior robustness and performance stability.

\begin{table}[]
\centering
\small
\begin{NiceTabular}{lcc}
\toprule
\textbf{Method} & \textbf{COMET} & \textbf{COMETKiwi} \\
\midrule
DPO & 36.86 & 22.61 \\
DPO (\textit{w.} SFT loss) & 80.69 & 75.60 \\
SimPO & 70.48 & 70.94 \\
ORPO & 80.23 & 74.94 \\
KTO & 80.30 & 75.04 \\
CPO (Standard) & 81.10 & 75.24 \\
\midrule
\textbf{Ours (StarPO)} & \bf 81.55 & \bf 77.15 \\
\bottomrule
\end{NiceTabular}
\caption{Comparison of different offline preference optimization algorithms.}
\label{table:offline_po_comparison}
\end{table}

\section{STAR Score Distribution}
\begin{figure}[]
    \centering
    \includegraphics[width=0.8\linewidth]{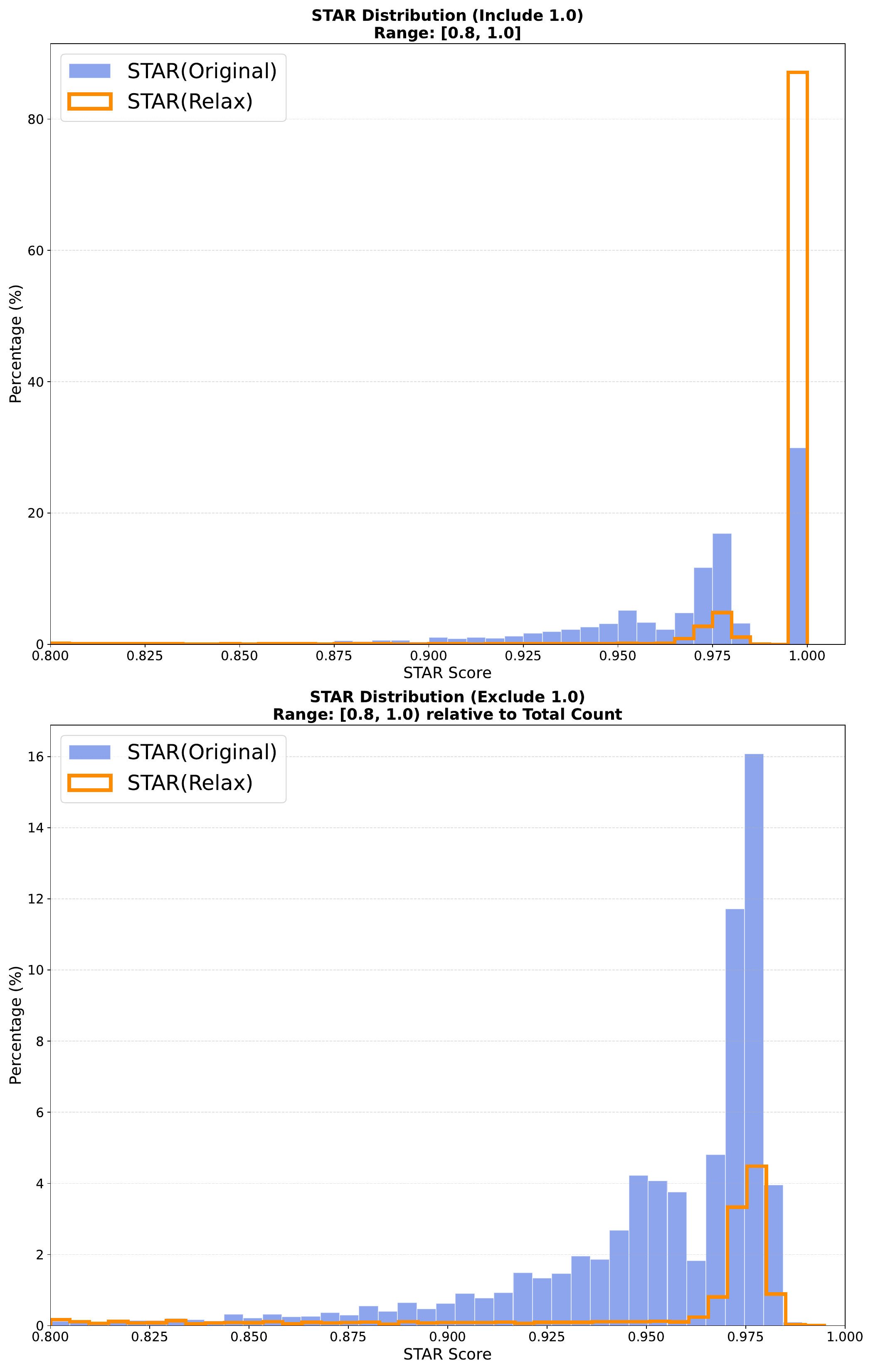}
    \caption{Histograms of metric scores for STAR (Original) and STAR (Relaxed). The \textbf{top} plot displays the full distribution including perfect matches (score=100 $\%$) The \textbf{bottom} plot zooms in by excluding perfect matches, highlighting that STAR (Original) maintains a dense distribution of high-quality candidates, whereas STAR (Relaxed) has sparse coverage in the near-perfect region.}    \label{fig:star_distribution}
\end{figure}

To further investigate the impact of structural constraints on preference pair construction, we visualize the score distributions of the standard STAR and its relaxed variant (STAR$_{\text{relax}}$) in Figure~\ref{fig:star_distribution}. 

As illustrated in the top panel, a vast majority of samples are clustered at the perfect score of 1.0 in the STAR (Relaxed) settings. The bottom panel of Figure~\ref{fig:star_distribution} provides a more granular view by excluding perfect 1.0 scores. Here, the contrast becomes more evident.

\label{apdx:star_distribution}

\section{Masked Token Distribution}
\label{apdx:mask_distribution}
\begin{figure}[]
    \centering
    \includegraphics[width=0.8\linewidth]{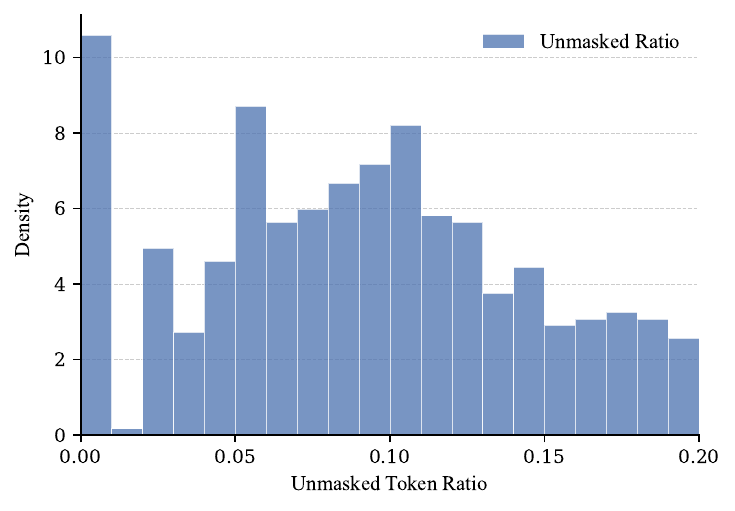}
    \caption{Frequency distribution of unmasked tokens. To provide a clear view of the token preservation statistics, the distribution is normalized}
    \label{fig:mask_distribution}
\end{figure}
Figure \ref{fig:mask_distribution} illustrates the probability density distribution of the unmasked token ratio.

\begin{figure*}[]
\centering
\includegraphics[width=\textwidth]{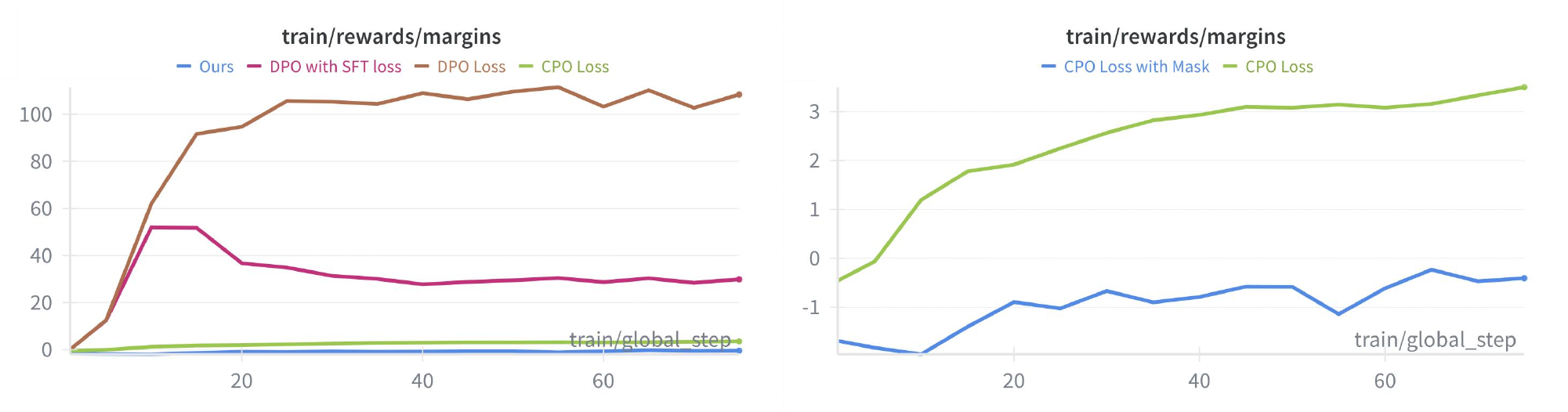}
\caption{Analysis of training margins. The \textbf{left} panel compares the margins of Ours, DPO with SFT loss, standard DPO loss, and standard CPO loss. The \textbf{right} panel provides a zoomed-in view focusing on the comparison between standard CPO loss and Ours (indicated as ``CPO Loss with Mask'' in the legend).}
\label{fig:margin}
\end{figure*}

\section{Theoretical Justification and In-Depth Analysis of Masking Strategies}
\label{apdx:masking_analysis}

Below, we provide a theoretical justification for the experimental observations above. We analyze how the masking mechanism affects the reward margins, the loss magnitude, and the gradient flow, specifically how it prevents the vanishing gradient problem often encountered in preference optimization.

\paragraph{1. Margin Scaling via Masking}

Let the standard log-likelihood margin between the preferred target $y_w$ and the dis-preferred target $y_l$ be denoted as $\Delta_{\text{full}}$:
\begin{equation}
\Delta_{\text{full}}(x, y_w, y_l) = \log \pi_\theta(y_w|x) - \log \pi_\theta(y_l|x).
\end{equation}

In the STAR-Masked objective, the likelihood is computed over a subset of sentences where the mask $\mathcal{M}(t_j)=1$. Let $\mathcal{S}_{\text{mask}} \subset \{1, \dots, n\}$ be the set of indices for sentences retained by the mask. The masked margin $\Delta_{\text{STAR}}$ is:
\begin{equation}
\begin{split}
\Delta_{\text{STAR}}(x, y_w, y_l) &= \log \pi_{\text{STAR}}(y_w|x) \\
&\quad - \log \pi_{\text{STAR}}(y_l|x).
\end{split}
\end{equation}

Since $\log \pi_{\text{STAR}}$ aggregates log-probabilities over a subset of tokens relative to the full document, the masked margin can be viewed as a scaled version of the full margin. Assuming the preference signal is distributed across the document, removing a portion of tokens (via $\mathcal{M}$) reduces the accumulated difference between $y_w$ and $y_l$. Specifically, if the mask retains a ratio $\rho \in (0, 1)$ of the effective information:
\begin{equation}
|\Delta_{\text{STAR}}| \approx \rho \cdot |\Delta_{\text{full}}| < |\Delta_{\text{full}}|.
\end{equation}
This derivation aligns with the empirical observation that reward margins decrease after introducing the mask, as shown in Figure \ref{fig:margin}. 

\paragraph{2. Impact on Initial Loss Magnitude}

The preference loss component in CPO is defined as:
\begin{equation}
\mathcal{L}(\Delta) = - \log \sigma(\beta \cdot \Delta),
\end{equation}
where $\Delta$ is the margin. The function $f(z) = -\log \sigma(z)$ is strictly monotonically decreasing. Assuming the model has a basic capability to distinguish $y_w$ from $y_l$ (i.e., $\Delta > 0$), the reduced margin caused by masking implies:
\begin{equation}
0 < \beta \Delta_{\text{STAR}} < \beta \Delta_{\text{full}}.
\end{equation}
Due to the monotonicity of the loss function:
\begin{equation}
-\log \sigma(\beta \Delta_{\text{STAR}}) > -\log \sigma(\beta \Delta_{\text{full}}).
\end{equation}
Thus, $\mathcal{L}_{\text{STAR-CPO}} > \mathcal{L}_{\text{CPO}}$ at the early stages of training. This theoretically confirms why the initial loss is higher and decays more slowly: the model perceives the "distance" between candidates as smaller, interpreting the optimization task as more difficult.

\paragraph{3. Gradient Saturation and Sustained Learning}

The efficacy of the optimization depends on the magnitude of the gradients. The gradient of the loss with respect to the model parameters $\theta$ is:
\begin{equation}
\nabla_\theta \mathcal{L} = \frac{\partial \mathcal{L}}{\partial \Delta} \cdot \nabla_\theta \Delta.
\end{equation}
We focus on the scalar coefficient $\frac{\partial \mathcal{L}}{\partial \Delta}$, which modulates the strength of the update. For the CPO loss:
\begin{equation}
\frac{\partial \mathcal{L}}{\partial \Delta} = \frac{\partial}{\partial \Delta} \big( -\log \sigma(\beta \Delta) \big) = -\beta \cdot \big( 1 - \sigma(\beta \Delta) \big).
\end{equation}
We compare the gradient coefficients in two scenarios:

\textbf{Scenario A: Full Objective (Standard CPO).}
If the model easily distinguishes $y_w$ from $y_l$ using simple patterns (e.g., trivial lexical differences in unmasked regions), $\Delta_{\text{full}}$ becomes large. As $\beta \Delta_{\text{full}} \to \infty$, $\sigma(\beta \Delta_{\text{full}}) \to 1$. Consequently, the gradient coefficient approaches zero:
\begin{equation}
| \nabla \mathcal{L}_{\text{CPO}} | \propto | 1 - \sigma(\beta \Delta_{\text{full}}) | \approx 0.
\end{equation}
This leads to gradient saturation, where the model stops learning effectively even if structural errors persist.

\textbf{Scenario B: Masked Objective (StarPO).}
By masking out easy-to-align sentences (or random segments), we force $\Delta_{\text{STAR}}$ to be smaller. The value of $\sigma(\beta \Delta_{\text{STAR}})$ stays further from 1 (closer to the linear regime of the sigmoid function).
\begin{equation}
| 1 - \sigma(\beta \Delta_{\text{STAR}}) | > | 1 - \sigma(\beta \Delta_{\text{full}}) |.
\end{equation}

Therefore, the masking mechanism acts as a regularizer that prevents the model from achieving a trivial margin on the training data. By artificially reducing the margin ($\Delta_{\text{STAR}}$), the objective ensures that the gradient magnitude remains significant throughout the training process. This explains why, despite a slower decrease in loss, the model performs better in later steps: it avoids early saturation and continues to optimize the policy on the complex, structurally critical segments represented by the unmasked tokens.
 \end{document}